\documentclass[review]{elsarticle}

\makeatletter
\def\ps@pprintTitle{%
  \let\@oddhead\@empty
  \let\@evenhead\@empty
  \let\@oddfoot\@empty
  \let\@evenfoot\@oddfoot
}
\makeatother

\usepackage{lineno,hyperref}

\usepackage{amsmath,amsfonts,bm}

\def\eqref#1{equation~\ref{#1}}

\def\1{\bm{1}}

\DeclareMathAlphabet{\mathsfit}{\encodingdefault}{\sfdefault}{m}{sl}
\SetMathAlphabet{\mathsfit}{bold}{\encodingdefault}{\sfdefault}{bx}{n}

\usepackage{hyperref}
\usepackage{url}

\usepackage{amsthm,amsmath,amsfonts,amssymb}
\newcommand{\norm}[1]{\left\lVert#1\right\rVert}

\usepackage{mathtools}
\usepackage{dirtytalk}

\usepackage{graphicx}
\usepackage{caption}
\usepackage{subcaption}
\usepackage{tabularx}
\usepackage{booktabs}
\usepackage{multirow}
\usepackage{float}

\newcolumntype{B}{>{\hsize=.53\hsize}X}
\newcolumntype{S}{>{\hsize=.47\hsize}X}

\modulolinenumbers[5]
\let\linenumbers\nolinenumbers\nolinenumbers 

\journal{arXiv} 

\begin{document}

\begin{frontmatter}

\title{Improving State-of-the-Art in One-Class Classification by Leveraging Unlabeled Data}


\author[jbr]{Farid Bagirov}

\author[jbr]{Dmitry Ivanov}

\author[jbr]{Aleksei Shpilman}


\address[jbr]{JetBrains Research}

\begin{abstract}

When dealing with binary classification of data with only one labeled class data scientists employ two main approaches, namely One-Class (OC) classification and Positive Unlabeled (PU) learning. The former only learns from labeled positive data, whereas the latter also utilizes unlabeled data to improve the overall performance. Since PU learning utilizes more data, we might be prone to think that when unlabeled data is available, the go-to algorithms should always come from the PU group. However, we find that this is not always the case if unlabeled data is unreliable, i.e. contains limited or biased latent negative data. We perform an extensive experimental study of a wide list of state-of-the-art OC and PU algorithms in various scenarios as far as unlabeled data reliability is concerned. Furthermore, we propose PU modifications of state-of-the-art OC algorithms that are robust to unreliable unlabeled data, as well as a guideline to similarly modify other OC algorithms. Our main practical recommendation is to use state-of-the-art PU algorithms when unlabeled data is reliable and to use the proposed modifications of state-of-the-art OC algorithms otherwise. Additionally, we outline procedures to distinguish the cases of reliable and unreliable unlabeled data using statistical tests.\footnote{The code is available at \url{https://github.com/jbr-ai-labs/PU-OC}}.

\end{abstract}


\end{frontmatter}

\linenumbers

\section{Introduction}

An input of a supervised binary classifier consists of two sets of examples: positive and negative. However, the access to labeled samples from both classes can be restricted in many realistic scenarios. A particularly well-studied restriction is the absence of labeled negative examples, which appears in malfunction detection, medical anomalies detection, etc. \cite{chalapathy2019deep}. One of the approaches that deal with this is One-Class (OC) classification \citep{moya1993one}. Typically, OC algorithms treat available positive examples as normal data and try to separate it from unseen data, that is often referred to as either anomalies or outliers \citep{grubbs1969procedures,hodge2004survey,chandola2009anomaly,chalapathy2018anomaly,chalapathy2019deep}. Further studies noted that OC algorithms always make assumptions about negative data distribution \citep{scott2009novelty,ruff2020unifying,ruff2020rethinking}. For example, some methods assume negative distribution to be uniform \citep{vert2006consistency,scott2006learning} or concentrated where positive data are rare \citep{tax2004support, ruff2018deep}, model negative distribution as a Gaussian \citep{oza2018one}, or separate positive data from the origin \citep{scholkopf2000support}. This may lead to errors if unlabeled data are not distributed according to these assumptions \cite{zhang2021understanding}.

\begin{figure*}
    \centering
    \captionsetup[subfigure]{justification=centering}

    \subfloat[][OC (no shift) \\ ROC AUC=0.92]{\includegraphics[width=0.24\textwidth]{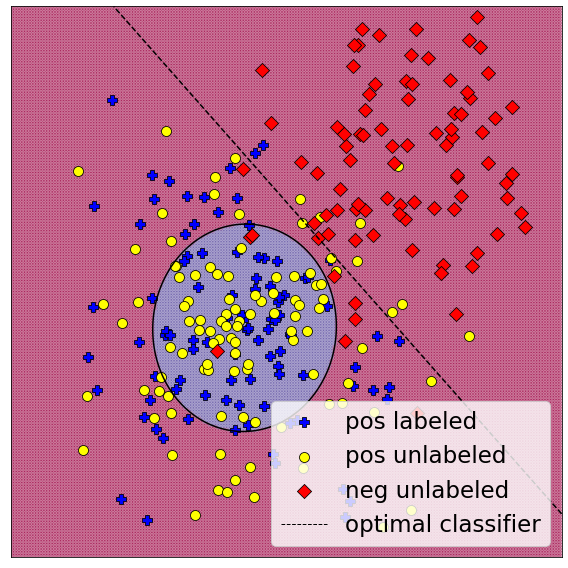}} 
    \subfloat[][PU (no shift)\\ ROC AUC=0.97]{\includegraphics[width=0.24\textwidth]{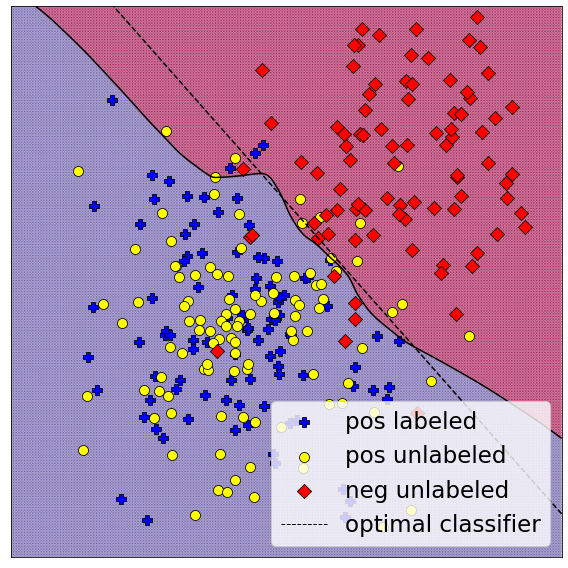}} 
    \subfloat[][OC (negative shift)\\ ROC AUC=0.91]{\includegraphics[width=0.24\textwidth]{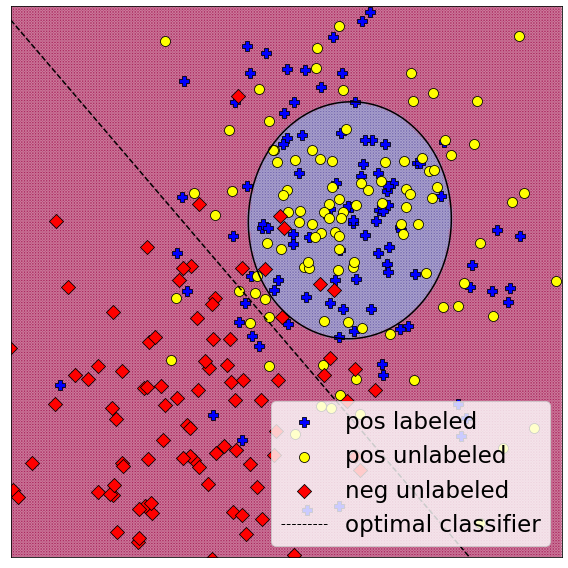}}
    \subfloat[][PU (negative shift)\\ ROC AUC=0.08]{\includegraphics[width=0.24\textwidth]{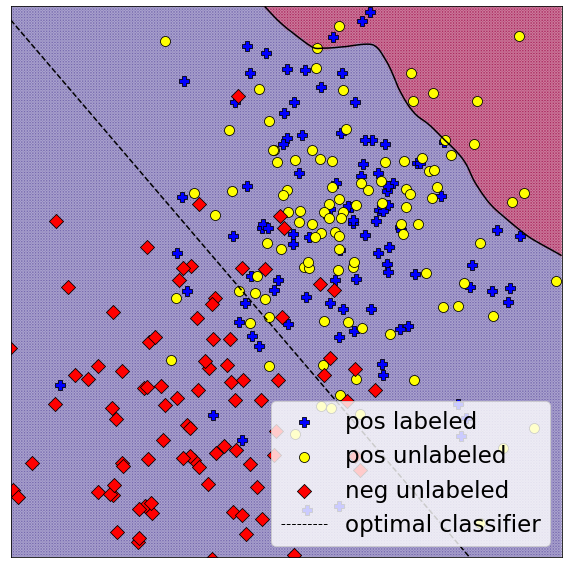}}

    \caption{Performance of OC and PU models on synthetic data. Subfigures (a) and (b) show the performance of the algorithms in the standard case when the negative distribution does not change from train to test times. Subfigures (c) and (d) show the effects of the negative distribution shift. The OC and PU models are OC-SVM and PU-SVM, respectively (section \ref{sec:methods_oc_svm}). The OC model is trained on labeled positive examples. The PU model is trained on labeled positive as well as unlabeled examples, i.e. a mixture of positive and negative examples. Black dashed lines represent the decision boundaries of the optimal Bayesian classifier. Overdependence on unreliable unlabeled data causes the PU method to misclassify all anomalies when a shift of the negative distribution occurs.}
    \label{synt1}
\end{figure*}

Another approach to classification in the absence of labeled negative data is Positive-Unlabeled (PU) learning \citep{denis1998pac,denis2005learning,li2005learning}. In addition to a labeled positive sample, PU algorithms leverage an unlabeled set of mixed positive and negative examples. In contrast to OC methods, PU methods do not make assumptions about negative data and instead approximate either negative distribution \citep{elkan2008learning,ivanov2019dedpul}, its statistics \citep{du2014analysis,du2015convex,kiryo2017positive}, or its samples \citep{Yu2002PEBL,liu2002spy,li2003rocchio,xu2019revisiting}, by comparing positive and unlabeled data. This approach can even outperform supervised classification, given a sufficient amount of unlabeled data \citep{niu2016theoretical,kiryo2017positive}.
Because PU methods make fewer assumptions and have access to more data, they might seem favorable to OC methods whenever unlabeled data are at hand. However, our experiments show that dependence on unlabeled data may hinder PU methods in particularly extreme cases, which we refer to as cases of unreliable unlabeled data. We identify several such cases, including distributional shifts in unlabeled data, scarcity of unlabeled data, and scarcity of latent negative examples in unlabeled data. We present a motivational example of a possible effect of distributional shifts on PU models in Figure \ref{synt1}. In subplots (a, b) a PU model approximates the separating line more accurately than an OC model. Conversely, in subplots (c, d) a shift of negative distribution causes the PU model to misclassify all negative examples, whereas the OC model is unaffected by the shift. While the presented example is synthetic, similar situations can occur in the real world, e.g. in the problem of spam detection where the spam bots constantly improve.

One possible conclusion is to opt for using OC methods when unlabeled data is unreliable, but we aim to find a way to construct robust methods that can learn even from unreliable unlabeled data. To this end, we propose modifications of several modern OC algorithms that leverage unlabeled data, which we refer to as PU-OC models. These modifications either are based on risk estimation techniques where the negative risk is approximated using positive and unlabeled data \citep{kiryo2017positive}, or simply replace positive data with unlabeled in algorithm-specific routines (see Sections \ref{sec:methods_oc_drocc}, \ref{sec:methods_oc_csi}). We find that all PU-OC methods benefit from reliable unlabeled data, but only modifications from the second group are safe to apply to unreliable unlabeled data (i.e. either improve upon or perform on par with the original OC methods). We pinpoint this result to a crucial property: in the absence of latent negative examples, the PU-OC algorithm becomes equivalent to the original OC algorithm. Our main practical recommendation is to use state-of-the-art PU algorithms when reliable unlabeled data is available and to use the PU-OC algorithms that satisfy this property when the robustness is a concern.

A question remains how to identify the cases of unreliable unlabeled data. We find that scarcity of latent negatives can be statistically tested by comparing predictions of OC or PU models for positive and unlabeled samples. Similarly, a shift of negative distribution can be tested by comparing predictions for unlabeled data from training and testing distributions, providing the latter is available. We provide a detailed procedure for this in subsection \ref{subsec:unreliable}. 

\paragraph{Related Work}

There is a line of work in the OC literature that investigates ways to augment OC methods with additional data. Several studies find that exposure to a small and possibly biased sample of outliers can improve the performance of OC classifiers \citep{hendrycks2018deep,ruff2019deep,ruff2020rethinking}. \cite{scott2009novelty} show both theoretically and empirically that unlabeled data help a classic machine learning algorithm to detect novelties. Nevertheless, there is a lack of modern literature that views OC and PU learning as different solutions to the same problem and gives practical recommendations for different scenarios. Our study attempts to fill this gap.

The problem of unreliable unlabeled data is also underexplored in PU literature. One existing direction is concerned with robustness to class prior shift \citep{charoenphakdee2019positive}. While this is a useful property, we study a more severe case of arbitrary negative distribution shifts. Furthermore, an increase in proportions of positive class can be seen as a special case of negative distribution shift. We implement the latest PU method robust to class prior shifts as a baseline \citep{nakajima2021drpu}. Other papers are concerned with arbitrary positive shifts \citep{hammoudeh2020learning} and covariate shifts \citep{sakai2019covariate}, but unlike us, they assume access to unbiased unlabeled data from the testing distribution. To the best of our knowledge, we address the problem of PU learning with shifted or scarce latent negatives for the first time.

Our study is conceptually similar to the field of safe semi-supervised learning where the focus is on making unlabeled data never hurt while retaining performance \citep{li2014towards,li2019safe,guo2020safe}. The difference is that we do not assume access to labeled negative data.

\section{Problem setup}\label{sec:problem_setup}

Let $x\in \mathbb{R}^d$ be a data point. Let $y \in \{0, 1\}$ be a binary label of $x$ that denotes its class. Let $s\in \{0, 1\}$ be a binary label of $x$ denotes whether $x$ is labeled, i.e. if $y$ is known. We view $x, y, s$ as random variables with some joint distribution $p(x, y, s)$. We assume that only positive data can be labeled, i.e. $p(s = 1 | y = 0) = 0$. The probability density functions of positive ($y=1$) and negative ($y=0$) distributions are given by:

\begin{subequations}
\bgroup
\def\arraystretch{0}
\begin{tabularx}{\textwidth}{Xp{0cm}X}
\begin{equation}
  p_p(x) \coloneqq p(x\mid y=1, s=0)
\end{equation}
& &
\begin{equation}
  p_n(x) \coloneqq p(x\mid y=0, s=0)
\end{equation}
\end{tabularx}
\egroup
\end{subequations}

Let $p_u(x)$ be the probability density function of the unlabeled distribution, i.e. a mixture of positive and negative distributions, and $\alpha = p(y=1|s=0)$ be the mixture proportion. We assume that the probability of being labeled $p(s = 1 \mid y = 1)$ is constant and independent of $x$, which is known as Selected Completely At Random (SCAR) setting \citep{elkan2008learning}. In this case:

\begin{gather}
    \label{unl_dist}
    p_u(x) \coloneqq p(x|s=0) =  \alpha p_p(x) + (1-\alpha) p_n(x)
\end{gather}

In OC methods we assume that only a labeled sample $X_p$ from $p_p(x)$ is available. In PU methods we assume the \textit{case-control scenario} \citep{bekker2020learning}: two sets of data $X_p$ and $X_u$ are sampled independently from $f_p(x)$ and $f_u(x)$. Both OC and PU methods output some score function proportional to $p(x)\coloneqq p(y=1|x, s=0)$ that separates positive and negative data. PU methods often additionally estimate $\alpha$. Since $\alpha$ is generally unidentifiable \citep{blanchard2010semi}, we will focus on estimation of its upper bound $\alpha^*$, as proposed in \citep{jain2016nonparametric}.

\section{Methods}\label{sec:methods}

In this section, we first describe the PU methods used in this study. We begin by describing Risk Estimation approach, which some of our PU-OC algorithms are based on. Then, we describe DRPU and PAN, which we use for comparison as the most recent and state-of-the-art PU algorithms. Finally, we describe the OC methods used in this study and propose their modifications that leverage unlabeled data, i.e. the PU-OC algorithms. In \ref{app:experiments_pu} and \ref{app:experiments_oc}, we verify our choices of the state-of-the-art for both OC and PU by comparing these algorithms with other modern algorithms. Additional details about the methods are reported in \ref{app:sec_methods}.

\subsection{Positive-Unlabeled methods}\label{sec:pu_methods}

\subsubsection{Risk Estimation}

Let $h(x)$ be an arbitrary decision function that estimates $y$, $l(t, y)$ be the loss incurred for predicting $t$ when the ground truth is $y$. Define $R^+_p(h)=E_{x\sim f_p}(l(h(x), 1))$ and $R^-_n(h)=E_{x\sim f_n}(l(h(x), 0))$ as the positive and the negative risks. If both positive and negative data are available, the risk of the decision function can be estimated as a weighted sum of positive and negative risks (eq. \ref{pnre}). In the PU setting,  $R^-_n(h)$ is unavailable but can be estimated as a difference between the risks on positive and unlabeled data (eq. \ref{upu}), as shown in \citep{du2014analysis, du2015convex}.

\begin{subequations}
\bgroup
\def\arraystretch{0}
\begin{tabularx}{\textwidth}{Sp{0cm}B}
\begin{equation}
\label{pnre}
  R_{pn}(h) = \alpha R^+_p(h) + (1 - \alpha)  R^-_n(h)
\end{equation}
& &
\begin{equation}
\label{upu}
  R_{pu}(h) = \alpha R^+_p(h) - \alpha  R^-_p(h) + R^-_u(h)
\end{equation}
\end{tabularx}
\egroup
\end{subequations}

Estimator (\ref{upu}) is called unbiased risk-estimator and can be improved by introducing a non-negativity constraint to reduce overfitting \citep{kiryo2017positive}:
    
    \begin{gather}
        \label{nnpu}
        R_{nn}(h) = \alpha R^+_p(h) + \max(0,  -\alpha  R^-_p(h) +  R^-_u(h))
    \end{gather}
    
Estimator (\ref{nnpu}) is called non-negative risk estimator. In practice, the decision function $h$ is parameterized by $\theta$, which can represent weights of a neural network or some other model. The parameters are trained to minimize $R_{nn}(h(x \mid \theta))$ for some loss function like double hinge \citep{du2015convex} or sigmoid \citep{kiryo2017positive}. We use the latter. Notice that $\alpha$ is assumed to be identified in this method, so in experiments, we additionally estimate it with TIcE \citep{bekker2018estimating}. 

Risk estimation can be applied to modify any OC model to leverage unlabeled data, providing this OC model is based on or can be generalized to the supervised (PN) setting. This can be done by first replacing the OC objective with the PN objective, and then applying risk estimation to the PN objective, i.e. evaluating negative risk using positive and unlabeled samples. 

\subsubsection{DRPU}
    Density Ratio estimation for PU Learning (DRPU) \citep{nakajima2021drpu} is a new state-of-the-art PU method that can be viewed as a combination of Risk Estimation techniques and Density Ratio estimation techniques. From the Bayes rule we have:
    
    \begin{gather}
        p(y=1|x, s=0) = \frac{p(x| y=1, s=0)p(y=1|s=0)}{p(x|s=0)} = \alpha \frac{p_p(x)}{p_u(x)}
        \label{pudre}
    \end{gather}
    
    Equation \ref{pudre} shows that PU learning can be decomposed into the problems of estimating $\alpha$ (Mixture Proportion estimation) and estimating the ratio $r(x)=\frac{p_p(x)}{p_u(x)}$, (Density Ratio estimation). DRPU tackles the latter problem via minimization of the Bregman divergence \citep{sugiyama2012density} and does not require the identified $\alpha$ during training. As a result, DRPU is stable to the shifts of class proportions in unlabeled data. After estimating $r(x)$, the class prior $\alpha$ can be approximated on the validation dataset as $\min\limits_{x_v} \frac{1}{r(x_v)}$. Note that misestimating the class prior only shifts the predictions while retaining their relative order, and therefore does not affect the ROC AUC score. In our study we adapt official implementation of DRPU \footnote{\href{ https://github.com/csnakajima/pu-learning}{\label{drpu}https://github.com/csnakajima/pu-learning}}.

\subsubsection{PAN}
    Predictive Adversarial Network (PAN) \citep{hu2021predictive} is another state-of-the-art PU algorithm based on a GANs \citep{goodfellow2014generative}. In this algorithm, two classifiers C and D play a minimax game instead, which differs from the generator and the discriminator trained in the standard GAN framework. The objective of the classifier C is to predict the same probabilities on the unlabeled data as the discriminator D. As for the discriminator D, its loss consists of two parts: the negated loss of the classifier C and the negated log-likelihood incurred by the classification of the positive data against the unlabeled. As a result, C tries to mimic D, while D tries to both confuse C and classify the given data correctly. In the equilibrium, the two models equivalently classify the unlabeled data while distinguishing the positive data from the negative. We use our implementation of PAN based on the original paper.


\subsection{One-Class methods}\label{sec:methods_oc}

\subsubsection{OC-SVM}\label{sec:methods_oc_svm}

OC-SVM \citep{scholkopf2000support} is one of the most well-known OC methods. OC-SVM modifies the classical objective of Support Vector Machines and tries to separate the labeled positive data from the origin. Similarly to SVM, OC-SVM can be optimized by solving the dual problem and specifying only dot product in the feature space, i.e. kernel trick. In the case of RBF kernel (and other translation-invariant kernels \citep{scholkopf2001estimating}), OC-SVM tries to envelope the positive data while minimizing the enclosed volume and is equivalent to SVDD \citep{david2001tax,tax2004support}. OC-SVM solves the following optimization problem:

    \begin{gather}
        \label{oc_svm}
        \min_{w, r} \frac 12 \norm{w}^2 -r+ \frac{1}{\nu N} \sum_{i=0}^N \max(0, r - w \cdot \Phi(x_i)) -r
    \end{gather}
    
where $\Phi$ is a feature map in some Hilbert space, $\nu$ is both a regularization parameter and a correction for possible data contamination, and $w \cdot \Phi(x_i) - r$ is the decision function. In the modern literature, OC-SVM is mainly used as a benchmark algorithm. However, new methods are still being proposed that are based on OC-SVM (\citep{ruff2018deep, chalapathy2018anomaly}) or incorporate it as a part of the model (\citep{ergen2017unsupervised, oza2018one}). We use scikit-learn implementation of OC-SVM in our work. Traditionally, OC-SVM is used with RBF kernel but in our implementation we use the linear kernel. During the preliminary experiments, we discovered that such replacement improves ROC AUC of OC-SVM on benchmark datasets. We train all SVM models on features extracted with an encoder, which was trained as a part of an autoencoder only on positive data.

\textbf{PU-OC}\quad As shown in \cite{du2015convex}, linear models can be trained in the PU setting by minimizing risk defined in (\ref{upu}). Since soft-margin SVM is a linear model that minimizes empirical risk with respect to hinge-loss (square brackets in \ref{sm_svm}) with regularization term,  one can use hinge loss as $h(x)$ and apply risk estimator techniques to soft-margin SVM. We refer to this approach as PU-SVM.
 
 \begin{gather}
        \label{sm_svm}
        \min_{w, b} \lambda \norm{w}^2 + \bigg[\frac 1N \sum\limits_{i=1}^N \max(0, 1 - y_i (\textbf{w}^t\textbf{x} - b))\bigg]
 \end{gather}
 
 In \ref{app:sec_methods_svm_dual}, we show that similarly to SVM and OC-SVM PU-SVM with unbiased risk estimator can be optimized by solving the dual problem, but that all labeled positive examples are support vectors in PU-SVM. Note that this is a secondary result that can be of independent interest. Since all labeled examples are considered as support vectors, such solution is computationally inefficient and does not scale well with data size. Instead, we apply non-negative risk (\ref{nnpu}) estimator for the SVM objective and solve it with Stochastic Gradient Decent (SGD) with $\textbf{w}$ as a parameter. Details of our optimization process are reported in \ref{app:sec_methods_svm_our}.

\subsubsection{OC-CNN}\label{sec:methods_oc_cnn}

OC-CNN \citep{oza2018one} is a hybrid model for one class classification that consist of two parts. First, the positive data is processed by a pretrained feature extractor (ResNet, VGG, AlexNet, etc.), which is frozen during both training and inference. After the latent representation of positive data is acquired, OC-CNN models the negative distribution as a Gaussian in the latent feature space. Then, a standard binary classifier like a multi-layered perceptron is trained to distinguish positive latent features from the pseudo-negative Gaussian noise. Additionally, the author claim that instead of the binary classifier, OC-SVM can be trained on latent representations, which also yields decent results. We implement OC-CNN according to the original paper. As in the original paper, we use ResNet pretrained on ImageNet as a feature extractor. Since pretraining is done in a supervised manner and classes from CIFAR10 overlap with classes from ImageNet, OC-CNN should not be directly compared with other OC algorithms that do not have access to such additional data. The same logic applies to the proposed PU modification.

\textbf{PU-OC} \quad Since OC-CNN assumes a particular negative distribution, it is easy to modify it for PU setup. We replace the pseudo-negative examples in the latent space with unlabeled examples processed through the same feature extractor and then perform the standard risk estimation procedure, i.e. minimize the non-negative risk in the latent space (\ref{nnpu}).

\subsubsection{OC-LSTM}

OC-LSTM \citep{ergen2017unsupervised} is a hybrid method that can work with text and series data. This model is based on a combination of the representation power of Long Short-Term Memory \citep{hochreiter1997long} and the OC-SVM objective. Specifically, the data points in a series are processed through an LSTM block and transformed to vector representations, e.g. as last hidden vector or average over all hidden vectors. After that, a single vector per series is acquired and fed into OC-SVM. Authors described two approaches to training. First, the updates of OC-SVM and LSTM can alternate: one model is frozen and the other is updated. Second, both parts can be updated simultaneously via gradient decent (same procedure for OC-SVM as in our PU-SVM optimization). We employ the second approach.

\textbf{PU-OC} \quad OC-LSTM with end-to-end optimization can be easily adapted to the PU setting. Since SVM is updated via SGD, we can replace it with our PU-SVM model and use same learning pipeline.

\subsubsection{DROCC}\label{sec:methods_oc_drocc}

DROCC \citep{goyal2020drocc} achieves the state-of-the-art performance on several real-world datasets across different domains. The key assumption behind DROCC is that the positive data lie on a low-dimensional locally Euclidean manifold. Based on this assumption, the pseudo-negative examples can be generated from the available positives. The negative examples are approximated with gradient ascent (similarly to adversarial attack) with some tricks. The ascent is performed from the positive point with the current decision function used as a measure of negativity. After that, the parameters of the decision function are optimized to minimize the standard binary classification loss between the positive and the pseudo-negative (adversarial) examples. We slightly modify the authors' implementation\footnote{\href{ https://github.com/microsoft/EdgeML}{https://github.com/microsoft/EdgeML}}. Originally, the adversarial search is performed in the input space. Because of that, negative examples are often original positive images with some noise. Instead, we search for negative examples in the feature space obtained from the output of the middle hidden layer of the network. In our experiments, we find that this small change improves ROC AUC. 

\textbf{PU-OC} A major bottleneck of DROCC is the generation of pseudo-negative examples. Since this search requires access to a decision function, it can struggle to find good approximation for negatives, especially early in the training. We propose to deal with this bottleneck using unlabeled data. In our PU-DROCC, the pseudo-negative points are generated from the unlabeled rather than the positive examples. This way, the pseudo-negative examples are closer to the real negatives if the unlabeled data contain some negative points, i.e. if $\alpha < 1$. In the extreme case when the unlabeled data contain only positives, our modification is equivalent to the original algorithm.

\subsubsection{CSI}\label{sec:methods_oc_csi}

CSI \citep{tack2020csi} is an OC method based on two self-supervised techniques: contrastive learning and transformation predictions. Contrastive learning works with a modified dataset where each data point is augmented with one of several transformations. The key idea is to move close the embeddings of different augmentations of the same initial objects and vice-versa move far the embeddings of different objects. In CSI some of augmentations (e.g. rotations) are also considered as negative examples. In addition to contrastive loss CSI is learned to predict applied transformations. The anomaly score of a data point consist of two parts: contrastive representation score and classification score. Contrastive representation score of the datapoint is the distance to the closest train point in embeddings' space, and classification score is an error in the prediction of transformations. Both scores should be low for positive sample. 
We use the official implementation of CSI \footnote{\href{ https://github.com/alinlab/CSI}{https://github.com/alinlab/CSI}} with a change in neural architecture. While original study uses ResNet, we train a small convolutional network instead for a fair comparison with other algorithms

\textbf{PU-OC}\quad In our PU modification of CSI, we additionally contrast positive points with examples from unlabeled data. Like our PU-DROCC, such modification will be close to the original model in the extreme case when unlabeled data has no negative examples, i.e. $\alpha\sim1$. 

\begin{table}[h]
    \caption{Datasets' details for different settings}
    \label{data_details}
    \centering
    \resizebox{\textwidth}{!}{
    \begin{tabular}{ccccccccr}
        \toprule
        Setting & Dataset & Pos class & Neg class & Pos lab & Pos unl & Neg unl & $\alpha$  \\\hline
        One-vs-all&CIFAR-10 & any class& \begin{tabular}[c]{@{}c@{}} all other \\ classes \end{tabular}&2500&2500&2500&0.5\\\hline
        Neg shift& CIFAR-10 & $\{0\}$ or $\{2\}$& \begin{tabular}[c]{@{}c@{}} rnd subset \\ from other \\ classes \end{tabular}&2500&2500&2500&0.5\\\hline
        Pos modes &CIFAR-10 &\begin{tabular}[c]{@{}c@{}} subset from \\ vehicle or \\ animal classes\end{tabular} & \begin{tabular}[c]{@{}c@{}} all other \\ classes \end{tabular} & 2500 & 2500 & 2500 & 0.5 \\\hline
        \begin{tabular}[c]{@{}c@{}}  Dependency \\ on $|X_u|$\end{tabular} &CIFAR-10&$\{0, 1, 8, 9\}$& \begin{tabular}[c]{@{}c@{}} all other \\ classes \end{tabular} & 2500 & vary & vary & 0.5 \\\hline
        \begin{tabular}[c]{@{}c@{}}  Dependency \\ on $\alpha$\end{tabular} &CIFAR-10 &$\{0, 1, 8, 9\}$& \begin{tabular}[c]{@{}c@{}} all other \\ classes \end{tabular} & 2500 & 2500 & 2500 & vary \\\hline
        --&Abnormal &Car&-- &576 & 573 &86  &0.87\\\hline
        --&PenDigits &--& --& 2533 &2509&110&0.97\\\hline
        --&Dec. Reviews &--& --&170&154&316& 0.33 \\\hline
        --&SMS Spam &--& --&1910&1937&609&0.76\\\hline
        --&Twitter & Genuine & Social1 bot & 423 & 424 &818 & 0.33\\
        \bottomrule
    \end{tabular}
    }
\end{table}

\section{Datasets}\label{sec:datasets}

Here we describe the datasets used in this study and some preprocessing details.

\paragraph{CIFAR-10}

CIFAR-10 is a standard for both OC and PU methods benchmark dataset with images from ten animal and vehicle classes \citep{krizhevsky2009learning}. There are around 6000 images for each class and the proportion of negatives and positives depends on the particular experiment.

\paragraph{Abnormal1001}

\begin{figure}
	
    \centering

        \includegraphics[width=.6\columnwidth]{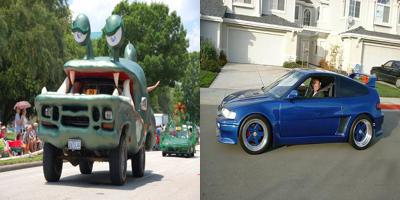}
        \captionof{figure}{Examples of abnormal car (left) and normal car (right) from Abnormal1001.}
        \label{abn_example}
\end{figure}

This dataset consists of abnormal images from six classes, including Chair, Car, Airplane, Boat, Sofa, and Motorbike \citep{saleh2013object}. Normal images come from the respective classes from the PASCAL VOC dataset \citep{everingham2010pascal}. An example of images from this dataset is presented in Figure \ref{abn_example}. This dataset is challenging for PU methods since the negative data are very scattered and only a few negative examples are available. We perform experiments only with Car class since it has the most examples, with 110 abnormal and 1315 normal images, $\alpha=0.87$. We use preprocessed Abnormal1001 dataset, where all images are resized to $200\times200$ resolution. Additionally for SVM-based models, we preprocess both CIFAR-10 and Abnormal1001 with encoders of autoencoders trained only on positive data. For CNN-based models, we apply standard normalization to feed images to ResNet.

\paragraph{Pendigits}

Pendigits dataset consists of consecutive pixels sampled from digits handwritten on a tablet \citep{asuncion2007uci}. The task is to classify the digits based on these sequences of pixels. We use the processed dataset from {ODDS}\footnote{http://odds.cs.stonybrook.edu} with 6870 examples in total and 156 anomalies, $\alpha=0.97$.

\paragraph{Text Datasets}  

We use two different datasets with text data. The first is standard dataset of deceptive reviews \citep{ott2011finding}. It consists of 400 deceptive hotel reviews acquired with crowdsourcing service and 400 truthful reviews from TripAdvisor, $\alpha=0.33$. The second is SMS Spam dataset \citep{asuncion2007uci}. It contains 5572 examples with 747 spam messages considered as anomalies, $\alpha=0.76$. In order to preprocess text datasets, we remove stop words and punctuation, cast characters to lower case, and apply Snowball Stemmer from the nltk package. Additionally, we use pretrained  {Google News embeddings}\footnote{https://code.google.com/archive/p/word2vec/}.

\paragraph{Twitter bots}

This dataset \citep{cresci2016dna, rodriguez2020one} provides information about around 7000 accounts, which are divided into genuine users and four different types of bots. On this dataset, we treat only one type of bot as available in the unlabeled sample during training. During inference, we have four different options for negative data: the remaining three bot classes and all bot classes together. We acquire the data from the repository\footnote{\href{ https://botometer.osome.iu.edu/bot-repository/datasets.html}{\label{twitter}https://botometer.osome.iu.edu/bot-repository/datasets.html}}, which has small differences in the number of classes and the account features with the dataset described in the original paper.

\paragraph{Train-test split} For CIFAR-10 we use standard train-test split. Other datasets we split randomly into train and test in proportion $4:1$.

\paragraph{Dataset details} Table \ref{data_details} presents dataset details for each experimental setting. OC methods use only positive labeled data. PU methods also use unlabeled data but do not know which is positive and which is negative.



\section{Experiments}\label{sec:experiments}

In this section, we describe experimental settings and report results. We repeat each experiment 10 times. Statistical significance is verified via paired Wilcoxon signed-rank test with a 0.05 P-value threshold. We indicate a significant difference in favor of PU-OC or OC method with by underlining the higher ROC AUC value. As noted before (Section \ref{sec:methods_oc_cnn}), the performance of OC-CNN and PU-CNN should not be directly compared to other models. The selected hyperparameters and tuning procedure are reported in \ref{app:hyperparameters}. Additional experiments are reported in \ref{app:experiments}.

\begin{table}[h]
    \caption{ROC AUC in one-vs-all setting}
    \label{ova}
    \centering
    \resizebox{\textwidth}{!}{
    \begin{tabular}{lcccccccccc|c}
        \toprule
        Pos class& 0    & 1    & 2    & 3    & 4    & 5    & 6   & 7    & 8    & 9   & avg \\\hline
        OC-SVM   & 0.78 & 0.67 & 0.60 & 0.67 & 0.63 & 0.64 & 0.74 & 0.66 & 0.68 & 0.75 & 0.68 \\
        PU-SVM   & \underline{0.79} & \underline{0.70} & \underline{0.62} & \underline{0.68} & \underline{0.65} & \underline{0.72} & \underline{0.74} & \underline{0.70} & \underline{0.77} & \underline{0.82} & \underline{0.72} \\\hline
     
        CSI    & 0.70 & 0.87 & 0.63 & 0.62 & 0.60 & \textbf{0.83 }& 0.73 & 0.80 & 0.86 & 0.89 & 0.76 \\
        PU-CSI & \underline{0.71} & \underline{\textbf{0.90}} & 0.61 & \underline{0.68} & 0.62 & \textbf{0.83} & \underline{0.79} & \underline{\textbf{0.92}} & \underline{\textbf{0.89}} & \underline{\textbf{0.91}} & \underline{0.79}\\\hline
        DROCC    & 0.76 & 0.74 & 0.64 & 0.61 & 0.71 & 0.67 & 0.75 & 0.71 & 0.78 & 0.79 & 0.72 \\
        PU-DROCC & \underline{0.80} & \underline{0.85} & \underline{0.75} & \underline{0.74} & \underline{0.76} & \underline{0.77} & \underline{0.81} & \underline{0.79} & \underline{0.84} & \underline{0.83} & \underline{0.79} \\\hline
        OC-CNN   & 0.74 & 0.80 & 0.64 & 0.69 & 0.77 & 0.76 & 0.81 & 0.73 & 0.78 & 0.83 & 0.76\\
        PU-CNN   & \underline{0.91} & \underline{0.94} & \underline{0.87} & \underline{0.86} & \underline{0.90} & \underline{0.90} & \underline{0.94} & \underline{0.92} & \underline{0.95} & \underline{0.93} & \underline{0.91}\\\hline
        DRPU   & \textbf{0.85} & 0.89 & \textbf{0.78} & \textbf{0.78} & \textbf{0.81} & 0.82 & \textbf{0.88} & 0.83 & 0.88 & 0.87 & \textbf{0.84}\\
        PAN   & \textbf{0.85} & 0.89 & 0.76 & 0.77 & 0.80 & 0.80 & \textbf{0.88} & 0.82 & 0.88 & 0.87 & 0.83\\
        \bottomrule
    \end{tabular}
    }
\end{table}

\subsection{One-vs-all}

A common experimental setting in OC papers for datasets with multiple classes is one-vs-all classification. In this setting, one class is treated as positive and all other classes constitute negative examples. We fix $\alpha$ in this setting at $0.5$, label half of the positive points, and conduct experiments with each class selected as positive. Note that since OC methods have access to only half of all positive points, their performance cannot be directly compared to the results reported in their original papers.

The ROC AUC metrics for OC and PU methods for all classes are presented in Table \ref{ova}. PU-OC algorithms consistently significantly outperform their OC counterparts with a few exceptions where the methods perform on par. Furthermore, DRPU and PAN outperform other algorithms on most classes, although on classes 7 through 9 our PU-CSI performs exceptionally well.

\begin{table}
	\begin{minipage}{0.47\linewidth}
        \captionof{table}{ROC AUC for different modalities of negative distribution under random negative shifts on CIFAR-10 with Plane (0) class chosen as positive}
        \label{shift}
        \resizebox{\textwidth}{!}{
        \begin{tabular}{lccccr}
            \toprule
            Neg modality & 1 & 2 & 3 & 4 \\
            \midrule
            OC-SVM    & \underline{\textbf{0.76}} & 0.78	& 0.80 & 0.748 \\ 
            PU-SVM      & 0.70 &	0.78 &	0.80 &	\underline{0.752}     \\ \hline
            CSI      & 0.68 &0.74 & 0.69 & 0.69 \\ 
            PU-CSI   & \underline{0.73} & \underline{0.76} & \underline{0.70} & 0.70 \\\hline
            DROCC    &0.73 &	0.74 & 0.73 & 0.72 \\ 
            PU-DROCC   & 0.73 &	\underline{0.77} &\underline{0.77}	&\underline{0.75}  \\\hline
            OC-CNN    & 0.72 &0.71 & 0.74 & 0.74 \\ 
            PU-CNN   & \underline{0.80} & \underline{0.78} & \underline{0.85} & \underline{0.83} \\\hline
            DRPU & 0.74 & 0.77 & 0.80 & \textbf{0.81}\\
            PAN & 0.74 & \textbf{0.79} & \textbf{0.81} & \textbf{0.81} \\
            \bottomrule
        \end{tabular}
        }
    \end{minipage}\hfill
	\begin{minipage}{0.47\linewidth}
		\captionof{table}{ROC AUC for different modalities of negative distribution under random negative shifts on CIFAR-10 with Bird (2) class chosen as positive}
        \label{bird_shift}
        \centering
        \resizebox{\textwidth}{!}{
        \begin{tabular}{lccccr}
            \toprule
            Neg modality & 1 & 2 & 3 & 4 \\
            \midrule
            OC-SVM   & 0.54 & 0.59 & 0.62 & 0.61 \\ 
            PU-SVM   & 0.54 & 0.61 & 0.64 & 0.60 \\ \hline
            CSI      & 0.59 & 0.61 & \underline{0.58} & 0.60 \\ 
            PU-CSI   & 0.54 & 0.59 & 0.56 & 0.57 \\\hline
            DROCC    & 0.69 & 0.68 & 0.64 & 0.64 \\ 
            PU-DROCC & \textbf{0.72} & \underline{\textbf{0.73}} & \underline{\textbf{0.71}} & \underline{0.70} \\\hline
            OC-CNN   & 0.63 & 0.58 & 0.58 & 0.64 \\ 
            PU-CNN   & 0.60 & \underline{0.73} & \underline{0.76} & \underline{0.81} \\\hline
            DRPU     & \textbf{0.72} & 0.70 & 0.70 & \textbf{0.72} \\
            PAN      & 0.64 & 0.65 & 0.70 & 0.69 \\
            \bottomrule
        \end{tabular}
        }
	\end{minipage}
\end{table}
\begin{table}[h]
    \caption{ROC AUC for particular shifts with Bird class (2) chosen as positive. We test shifts for unimodal and multimodal negative distribution. We consider following multimodal negative distributions: $A_3=\{3, 4, 5\}$, $A_4=\{6, 7\}$, $V_2=\{0, 1\}$, $V_3=\{8, 9\}$}
    \label{shift_anima}
    \centering
    \resizebox{\textwidth}{!}{
    \begin{tabular}{lcccccccc}
        \toprule
        Shift      & $0\rightarrow3$ & $3\rightarrow0$ & $0\rightarrow1$ & $3\rightarrow4$ & $V_2\rightarrow A_3$ & $A_3\rightarrow V_2$ & $V_2\rightarrow V_3$ & $A_3\rightarrow A_4$\\\hline
        OC-SVM     & 0.53 & \underline{\textbf{0.77}} & 0.66 & 0.47 & \underline{0.75} & 0.47 & \underline{0.52} & \underline{0.72} \\
        PU-SVM     & \underline{0.68} & 0.59 & \underline{0.72} & 0.47 & 0.68 & \underline{0.57} & 0.60 & 0.66 \\\hline
        CSI      & \textbf{0.70} & 0.52 & 0.66 & \textbf{0.63} & 0.66 & 0.59 & 0.51 & 0.59\\
        PU-CSI   & 0.69 & 0.55 & \underline{0.71} & 0.60 & \underline{0.69} & 0.60 & 0.50 & 0.58 \\\hline
        DROCC      & \underline{0.67} & 0.70 & 0.78 & 0.53 & 0.73 & 0.63 & 0.60 & 0.72 \\
        PU-DROCC   & 0.63 & 0.69 & 0.80 & \underline{0.58} & \underline{\textbf{0.88}} & \underline{\textbf{0.67}} & 0.59 & 0.68 \\\hline
        OC-CNN     & \underline{0.61} & \underline{0.77} & 0.66 & 0.55 & 0.78 & 0.56 & 0.57 & 0.71 \\
        PU-CNN     & 0.58 & 0.58 & \underline{0.83} & \underline{0.64} & \underline{0.97} & \underline{0.79} & 0.58 & 0.71 \\\hline
        DRPU      & 0.53 & 0.54 & \textbf{0.89} & 0.54 & 0.51 & 0.62 & \textbf{0.91}& \textbf{0.73} \\
        PAN       & 0.53 & 0.54 & 0.83 & 0.54 & 0.52 & 0.62 & \textbf{0.91} & \textbf{0.73} \\
        \bottomrule
    \end{tabular}
    }
\end{table}

\subsection{Shift of the negative distribution}

While in the previous setting PU algorithms have shined, that setting is convenient in that unlabeled data contain reliable latent negative examples. In the next setting, we investigate the effect of negative distribution shift, i.e. when latent negative examples are sampled from different distributions at train and test times. For example, this can be relevant if the negative distribution constantly shifts over time and the model cannot be retrained at each time step, or if a large but biased unlabeled sample is available for training, whereas an unlabeled sample used for inference is either too small or for some reason unavailable. Since negative shift only affects unlabeled data, only PU methods are expected to be sensitive to it (Fig. \ref{synt1}). Further, we suspect that the effect of negative shift on PU methods might decrease with the increased modality of the negative distribution due to negative data covering more latent space, so we additionally vary the number of negative classes.

For this setting, we consider $0$ (airplane) or $2$ (bird) class from CIFAR-10 as positive, $n$ random classes as negative train, and $n$ other random classes as negative test, where $n\in\{1, 2, 3, 4\}$. The results are presented in Table \ref{shift} and Table \ref{bird_shift}. Despite the distribution shifts that negatively affect PU methods, most PU-OC algorithms either significantly outperform or perform on par with their OC counterparts. The exception is OC-SVM, which outperforms all other algorithms when $n=1$. Evidently, even incorrect but data-driven estimates of the negative distribution used by PU algorithm are on average more helpful than the generic assumptions made by OC algorithms. Furthermore, for state-of-the-art PU algorithms increasing the number of negative classes partially mitigates the effect of negative shifts.

Although PU-OC models outperform their OC counterparts when the shift is random, there are particular shifts that can significantly harm the performance of PU and PU-OC algorithms. We present results of the experiments with non-random negative shifts, such as animal-animal, animal-vehicle, vehicle-animal, and vehicle-vehicle, in Table \ref{shift_nonrnd} and Table \ref{shift_anima}. The results vary from algorithm to algorithm. OC-SVM and in some cases OC-CNN outperform their PU-OC counterparts, highlighting that modifications based on risk estimation are not robust to negative distribution shifts. PU-CSI shows the same performance as regular CSI. On the other hand, PU-DROCC significantly outperforms its OC counterpart in most cases and at least performs on par in other cases. It also outperforms both state-of-the-art PU algorithms in all cases but one. Indeed, this algorithm shows the highest robustness to shifts in the latent negative distribution.

\begin{figure}[h]
  \centering
  \subfloat[][OC \\ ROC AUC = 0.07]{\includegraphics[width=.4\textwidth]{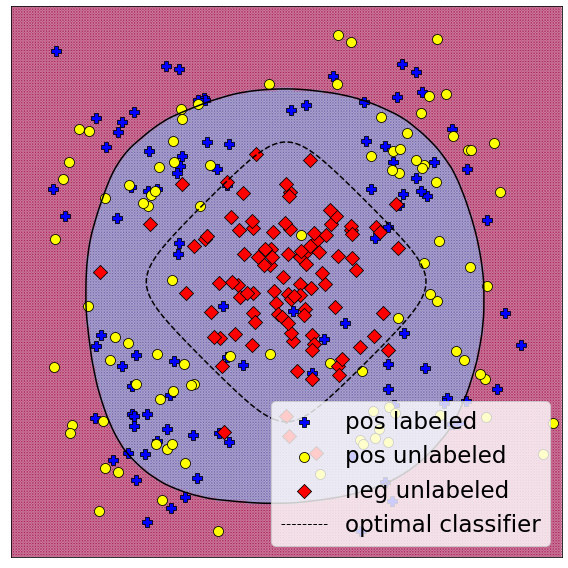}} 
    \subfloat[][PU \\ ROC AUC = 0.93]{\includegraphics[width=.4\textwidth]{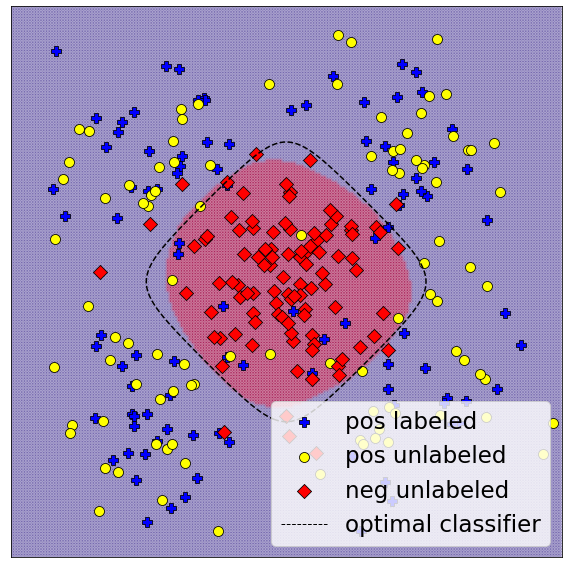}}
  \captionof{figure}{Separating lines of OC and PU algorithms on multimodal synthetic data. Positive distribution is a mixture of Gaussians with four different centers. The OC and PU models are OC-SVM and PU-SVM, respectively. OC-SVM tries to enclose all modes in a single envelope, whereas PU-SVM correctly identifies the location of negative data.}
  \label{synt2}
\end{figure}

\subsection{Number of positive modes}\label{app:experiments_modes}

In the standard one-vs-all setting the positive distribution is close to unimodal. Many OC methods try to envelope positive data and thus greatly benefit from unimodality \citep{ghafoori2020deep}. However, distributions of the real-world data are often more complex. Therefore, the one-vs-all setting might produce overly optimistic estimates of the performance of OC methods. We present a synthetic example that highlights the potential struggle of OC methods with multimodal positive distributions in the case when the negative points are concentrated between the modes of the positive distribution (Fig. \ref{synt2} a). In this example, the OC method attempts to enclose all positive modes in a single envelope and assigns high probability of being positive to negative examples. In contrast, PU methods are robust to multimodality and can accurately estimate the decision boundary (Fig. \ref{synt2} b).

In this setting, we consider a random subset from vehicle (animal) CIFAR-10 classes as positive data and all animal (vehicle) classes as negative data. We study the performance of models with respect to the number classes that form a positive sample. The results are reported in Tables \ref{cntrs}, \ref{cntrs2}. Like in the one-vs-all setting, PU-OC methods consistently outperform their OC counterparts. Further, the performance of all OC methods decreases with the increased number of modes. OC-SVM suffers from multimodality the most, which is expected from a method that tries to enclose all positive data in a single envelope. The performance of some PU-OC algorithms also drops due to data becoming more complex, but these drops are slight compared to OC algorithms. The exception is PU-CSI that performs exactly as well as the original CSI on this task. Interestingly, the performance of SOTA PU methods generally increases with the modality. As a result, DRPU and PAN outperform all other algorithms when the modality is high (as in other experiments, we exclude CNN-based models from this comparison due to leveraging a ResNet pretrained on ImageNet). However, when the modality is low, PU-DROCC achieves even better performance.

\begin{table}[h]
    \caption{ROC AUC for particular shifts with Airplane class (0) chosen as positive. We test shifts for unimodal and multimodal negative distributions. We consider following multimodal negative distributions: $A_1=\{2, 3, 4\}$,  $A_2=\{5, 6, 7\}$, and  $V_1=\{1, 8, 9\}$}
    \label{shift_mecha}
    \centering
    \resizebox{\textwidth}{!}{
    \begin{tabular}{lccccccc}
        \toprule
        Shift      & $1\rightarrow2$ & $2\rightarrow1$ & $1\rightarrow8$ & $2\rightarrow3$ &$V_1\rightarrow A_1$ & $A_1\rightarrow V_1$ & $A_1\rightarrow A_2$\\\hline
        OC-SVM   & \underline{\textbf{0.80}} & \underline{0.76} & 0.51 & \underline{\textbf{0.86}} & \underline{0.84} & \underline{0.64} & \underline{0.86} \\
        PU-SVM   & 0.77 & 0.63 & \underline{0.55} & 0.79 & 0.78 & 0.57 & 0.85 \\\hline

        CSI    & 0.75 & 0.54 & 0.46 & 0.54 & 0.75 & 0.49 & 0.80 \\
        PU-CSI & 0.76 & 0.56 & 0.47 & \underline{0.56} & 0.74 & 0.49 & 0.79\\\hline
        DROCC    & 0.74 & 0.76 & 0.60 & 0.80 & 0.75 & 0.68 & 0.79 \\
        PU-DROCC & 0.75 & \textbf{0.79} & \underline{0.66} & \underline{0.85} & \underline{0.77} & \textbf{0.69}& \underline{\textbf{0.87}} \\\hline
        OC-CNN   & \underline{0.80} & 0.60 & 0.50 & 0.80 & \underline{0.81} & 0.56 & 0.82 \\
        PU-CNN   & 0.63 & \underline{0.66} & \underline{0.65} & \underline{0.92} & 0.69 & \underline{0.71} & \underline{0.96} \\\hline
        DRPU    & 0.63 & 0.59 & \textbf{0.67} & 0.73 & 0.92 & 0.59 & 0.61 \\
        PAN    & 0.61 & 0.58 & 0.66 & 0.72 & \textbf{0.93} & 0.62 & 0.63 \\
        \bottomrule
    \end{tabular}
    }
    \label{shift_nonrnd}
\end{table}

\begin{table}[h]
        \caption{ROC AUC for different modalities of positive distribution on CIFAR-10 with vehicle classes chosen as positive}
        \label{cntrs}
        \centering
        \begin{tabular}{lccccr}
            \toprule
            Pos modality & 1 & 2 & 3 & 4 \\
            \midrule
            OC-SVM    & 0.74&	0.60&	0.58&	0.57 \\ 
            PU-SVM       &  \underline{0.80}&	\underline{0.80}&	\underline{0.77}&	\underline{0.80}      \\ \hline
            CSI    & 0.85&	0.82&	0.83	&0.88\\ 
            PU-CSI   &  0.85&	0.82&	0.83&	0.88\\\hline 
            DROCC    & 0.78&	0.79&	0.77&	0.73 \\ 
            PU-DROCC   &  \underline{\textbf{0.92}}&	\underline{\textbf{0.87}}&	\underline{0.86}&	\underline{0.83} \\\hline
            OC-CNN    & 0.90&	0.90&	0.81&	0.82 \\ 
            PU-CNN   &  \underline{0.98}&	\underline{0.98}&	\underline{0.96}&	\underline{0.96} \\\hline
            DRPU & 0.88 & 0.86 & \textbf{0.88} & \textbf{0.93} \\
            PAN  & 0.87 & 0.85 & \textbf{0.88} & \textbf{0.93} \\
            \bottomrule
        \end{tabular}
\end{table}

\begin{table}[h]
    \caption{ROC AUC for different modalities of positive distribution on CIFAR-10 with animal classes chosen as positive}
    \label{cntrs2}
    \centering
    \begin{tabular}{lcccccc}
        \toprule
        Pos modality& 1    & 2    & 3    & 4    & 5    & 6 \\\hline
        OC-SVM    & 0.60 & 0.64	& 0.69 & 0.66 & 0.64	& 0.67  \\
        PU-SVM    & \underline{0.63} & \underline{0.65}	& \underline{0.72} & \underline{0.68} & \underline{0.66}	& \underline{0.69}  \\\hline
        CSI      & 0.64 & 0.60 & 0.59 & 0.55 & 0.53 & 0.51 \\
        PU-CSI   & 0.66 & 0.60 & 0.59 & 0.53 & 0.52 & 0.51\\\hline
        DROCC     & 0.73 & 0.75 & 0.74 & 0.69 & 0.72 & 0.69 \\
        PU-DROCC  & \underline{\textbf{0.90}} & \underline{\textbf{0.91}}	& \underline{\textbf{0.90}} & \underline{\textbf{0.89}} & \underline{\textbf{0.89}} & \underline{0.90}\\\hline
        OC-CNN    & 0.76 & 0.81	& 0.70 & 0.78 & 0.74 & 0.74  \\
        PU-CNN    & \underline{0.97} & \underline{0.97}	& \underline{0.95} & \underline{0.97} & \underline{0.96} & \underline{0.96}  \\\hline
        DRPU     & 0.82 & 0.78 & 0.77 & 0.82 & 0.86 & \textbf{0.93} \\
        PAN      & 0.82 & 0.77 & 0.77 & 0.81 & 0.85 & \textbf{0.93} \\

        \bottomrule
    \end{tabular}
\end{table}

\subsection{Size and contamination of unlabeled data}

\begin{figure}[t]
    \centering
    \captionsetup[subfigure]{justification=centering}

    \subfloat[]{\includegraphics[width=.5\textwidth]{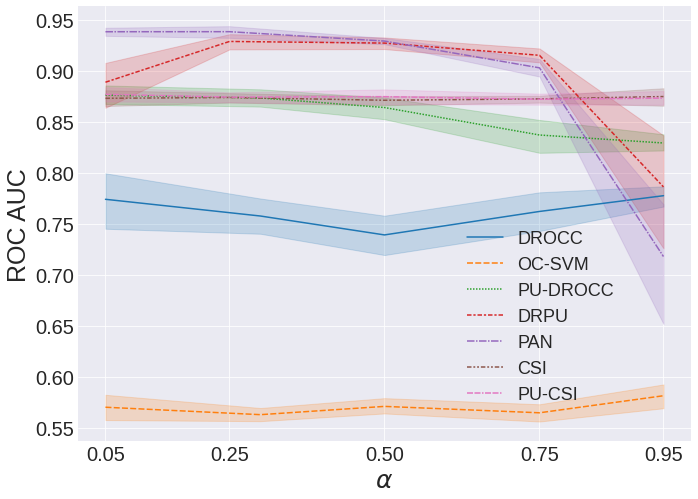}} 
    \subfloat[]{\includegraphics[width=.5\textwidth]{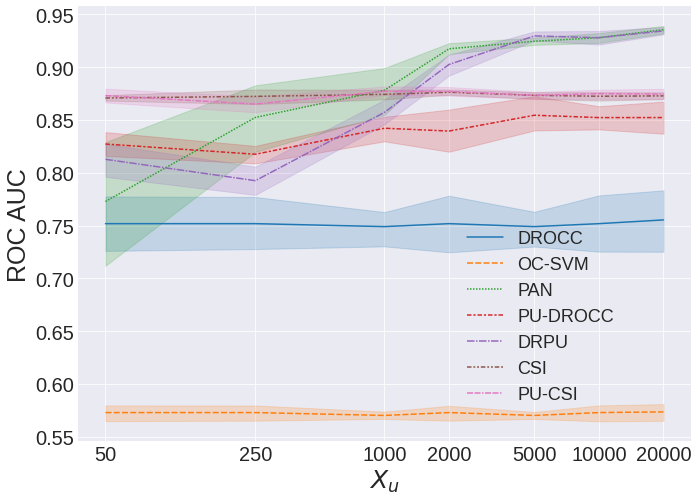}}
\caption{ROC AUC with respect to the proportion of positive examples in unlabeled data (a) and the size of unlabeled data (b).}
\label{unl}
\end{figure}
\begin{figure*}[h]
        \centering
        \begin{subfigure}[b]{0.475\textwidth}
            \centering
            \includegraphics[width=\textwidth]{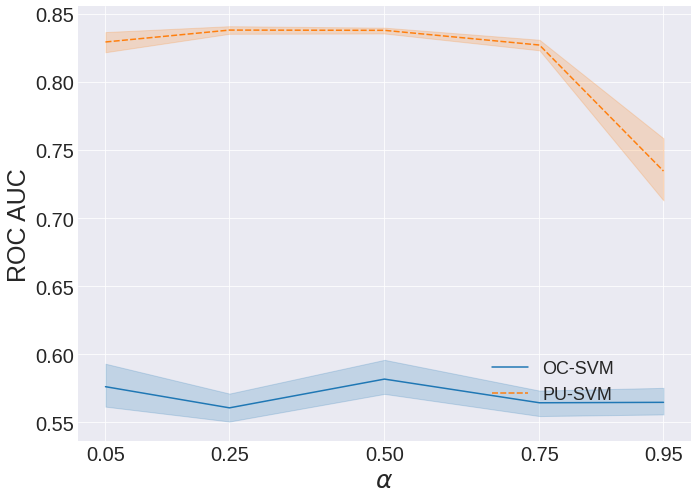}
            \caption[]%
            {{\small SVM-based models.}}    
        \end{subfigure}
        \hfill
        \begin{subfigure}[b]{0.475\textwidth}  
            \centering 
            \includegraphics[width=\textwidth]{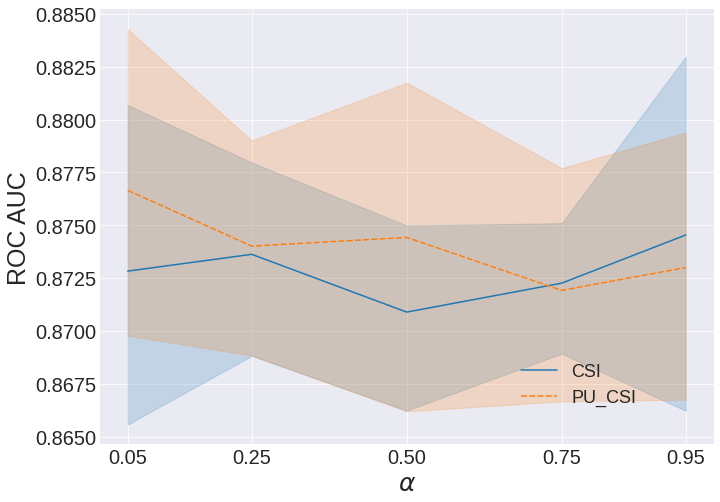}
            \caption[]%
            {{\small CSI-based models.}}    
        \end{subfigure}
        \vskip\baselineskip
        \begin{subfigure}[b]{0.475\textwidth}   
            \centering 
            \includegraphics[width=\textwidth]{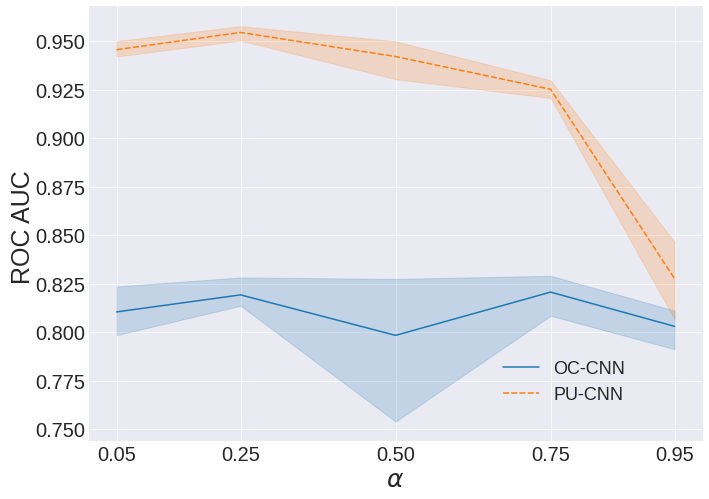}
            \caption[]%
            {{\small CNN-based models.}}    
        \end{subfigure}
        \hfill
        \begin{subfigure}[b]{0.475\textwidth}   
            \centering 
            \includegraphics[width=\textwidth]{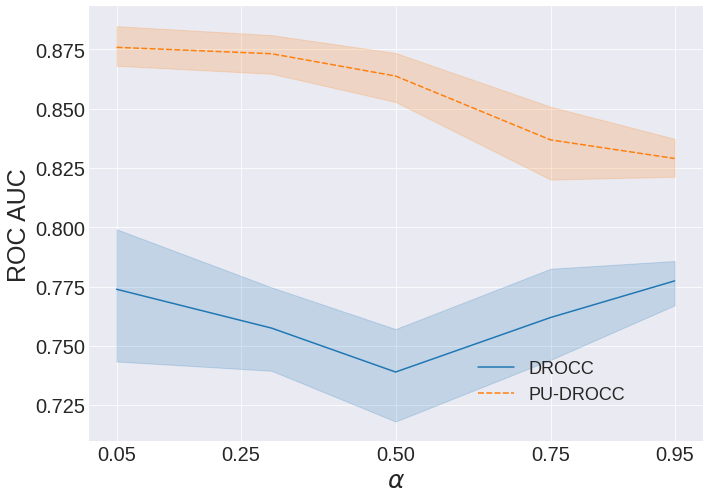}
            \caption[]%
            {{\small DROCC-based models.}}    
        \end{subfigure}
        \caption[ ]
        {\small ROC AUC with respect to proportion of positive examples in unlabeled data.} 
        \label{scu_alpha}
\end{figure*}
\begin{figure*}[h]
    \centering
    \begin{subfigure}[b]{0.475\textwidth}
        \centering
        \includegraphics[width=\textwidth]{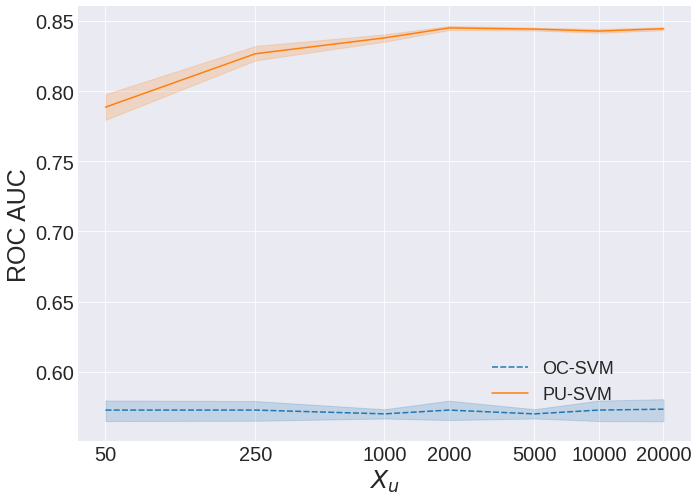}
        \caption[]%
        {{\small SVM-based models.}}    
    \end{subfigure}
    \hfill
    \begin{subfigure}[b]{0.475\textwidth}  
        \centering 
        \includegraphics[width=\textwidth]{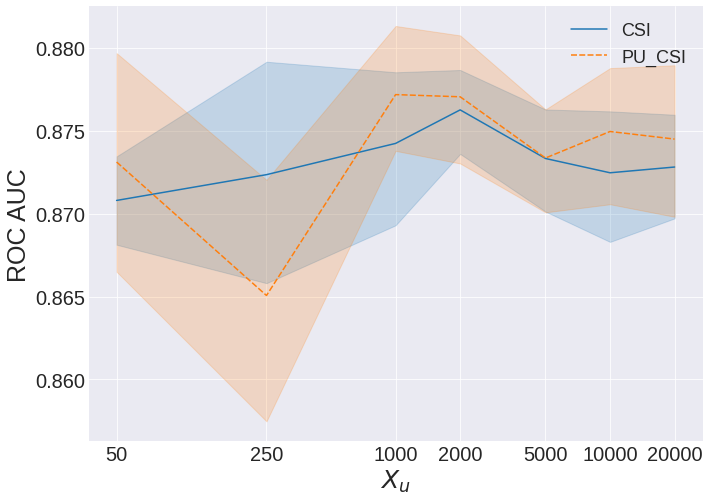}
        \caption[]%
        {{\small CSI-based models.}}    
    \end{subfigure}
    \vskip\baselineskip
    \begin{subfigure}[b]{0.475\textwidth}   
        \centering 
        \includegraphics[width=\textwidth]{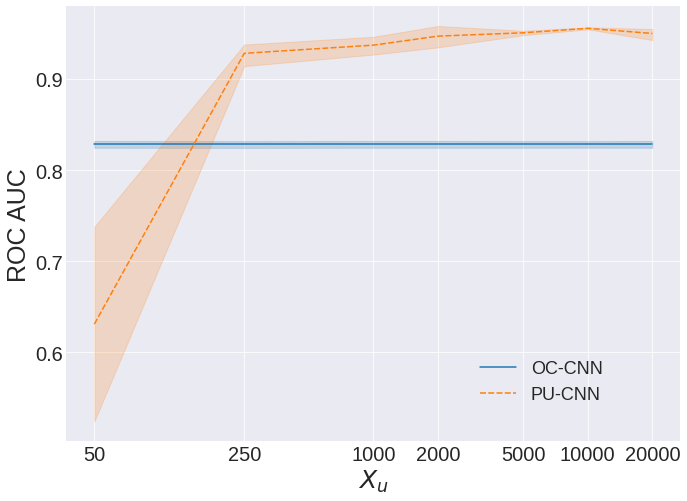}
        \caption[]%
        {{\small CNN-based models.}}    
    \end{subfigure}
    \hfill
    \begin{subfigure}[b]{0.475\textwidth}   
        \centering 
        \includegraphics[width=\textwidth]{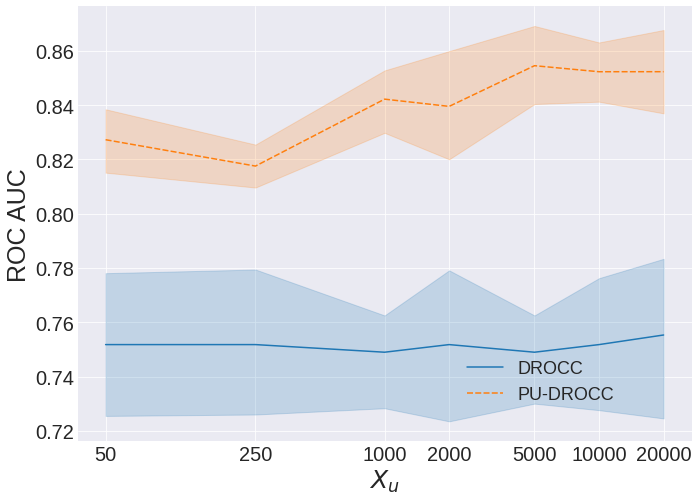}
        \caption[]%
        {{\small DROCC-based models.}}    
    \end{subfigure}
    \caption[ ]
    {\small ROC AUC for different sizes of available unlabeled data.} 
    \label{scu_size}
\end{figure*}

Other cases when PU algorithms may struggle are when negative examples are scarce, i.e. when the positive class proportion is high or the sample size of unlabeled data is small. We study the performance of OC and PU methods with respect to $\alpha$ and the size of the unlabeled sample. We mainly focus on the comparison of the state-of-the-art algorithms as well as some baselines . We choose all vehicles as positive classes and all animals as negative classes. The results are presented in Figure \ref{unl}. Subplot (b) shows that even as few as 50 unlabeled examples are sufficient for PU models to outperform DROCC, but that both versions of CSI outperform other algorithms until the unlabeled sample becomes bigger than 1000 examples. Subplot (a) shows that state-of-the-art PU methods struggle when unlabeled examples are mostly positive. In contrast, PU-DROCC consistently outperforms its OC analog. While both versions of CSI perform similarly, they outperform all other algorithms when $\alpha$ is high. Note that when $\alpha=0.05$, the difference between CSI and PU-CSI is statistically significant in favor of the latter. Also note that CSI generally performs much better than in the previous settings, as its performance varies with the selected positive classes. 
Additionally, we separately compare OC models with their PU-OC variants in this setting. Results are presented in Figure \ref{scu_alpha} and Figure \ref{scu_size}. Figure \ref{scu_alpha} shows that PU-OC models usually outperform original models or perform on par, even when $\alpha$ is particularly high. The result, however, is compounded with multimodality of positive distribution, which might have a negative effect on some OC models. Figure \ref{scu_size} shows that for most models as few unlabeled examples as 250 are enough to significantly improve performance.

\subsection{Abnormal1001}

The results of the experiments on Abnormal are presented in Table \ref{abnormal}. Although this dataset is challenging for PU methods due to the scarcity of anomalies and similarity of nominal and anomalous data, PU-OC models still outperform their OC counterparts across all models. Furthermore, PAN and our PU-DROCC perform the best among all methods.

\begin{table}
    \begin{minipage}{.4\linewidth}
        \caption{ROC AUC on Abnormal1001}
        \label{abnormal}
        \centering
        \begin{tabular}{lccr}
            \toprule
            Base & OC & PU \\
            \midrule
            SVM    &  0.60 & \underline{0.64}\\
            CSI    & 0.55 & \underline{0.64} \\
            DROCC  & 0.60 & \underline{\textbf{0.84}}\\
            CNN    & 0.53 & \underline{0.76}\\
            \hline
            DRPU & --& 0.78 \\
            PAN & --& \textbf{0.84} \\
            \bottomrule
        \end{tabular}
    \end{minipage}
    \begin{minipage}{.5\linewidth}
        \caption{ROC AUC on sequential data}
        \label{lstm}
        \centering
        \resizebox{\textwidth}{!}{

        \begin{tabular}{lcccr}
                    \toprule
                    Model & PenDigits & Deceptive & SMS Spam \\
                    \midrule
                    OC-LSTM    & 0.96& 0.52& 0.69 \\
                    PU-LSTM   &\textbf{ 0.97} & \underline{\textbf{0.77}} & \underline{\textbf{0.92}}  \\
                    \bottomrule
        \end{tabular}
        }
    \end{minipage}
\end{table}

\subsection{Sequential data}

Comparison of LSTM-based models on datasets with sequential data can be found in Table \ref{lstm}. PU models achieve better scores than OC algorithms on all three datasets. Furthermore, OC-LSTM performs poorly on the Deceptive Reviews dataset, which has very few labeled positive examples.

\subsection{Twitter bots}\label{app:experiments_twitter}

Results for experiments on Twitter dataset shown in Table \ref{twitter_res}. Since the dataset is tabular, it is difficult to apply CNN-based and CSI-based models, so we only apply the rest. In general, we find that PU-OC models outperform original models, especially in a more realistic case when new bots are added to the already existing. In most cases, state-of-the-art PU models achieve the best performance.

It is worth noting that the results of \citep{rodriguez2020one} for OC-SVM model differ from ours for some classes. We have discovered that with some seeds it is possible to replicate results of the original paper, and that the authors appear to have used one random seed in their experiments. However, when results are averaged by seed, the quality of OC-SVM model can drop significantly. Another possible explanation is the slight differences in the datasets that we and \cite{rodriguez2020one} use.

\begin{table}[h]
    \caption{ROC AUC for different negative test distributions on Twitter dataset.}
    \label{twitter_res}
    \centering
    \begin{tabular}{lccccc}
        \toprule
        Negatives& all    & social1    & social2    & social3    & traditional1\\\hline
        OC-SVM    & 0.61 & \underline{0.98}	& 0.57 & \underline{0.90} & 0.20  \\
        PU-SVM    & \underline{0.86} & 0.71 & \underline{0.89} & 0.77 & \underline{\textbf{0.98}}  \\\hline
        DROCC     & 0.72 & 0.93 & 0.74 & 0.95 & 0.28 \\
        PU-DROCC  & \underline{0.81} & \underline{0.96} & \underline{0.88} & 0.96 & 0.28\\\hline
        DRPU      & \textbf{0.94} & 0.94 & \textbf{0.96} & 0.93 & 0.89 \\
        PAN       & 0.83 & \textbf{0.99} & 0.87 & \textbf{0.99} & 0.45 \\
        \bottomrule
    \end{tabular}
\end{table}

\subsection{Identifying unreliable unlabeled data}\label{subsec:unreliable}

\begin{figure}[t]
    \centering
    \captionsetup[subfigure]{justification=centering}

    \subfloat[]{\includegraphics[width=.5\textwidth]{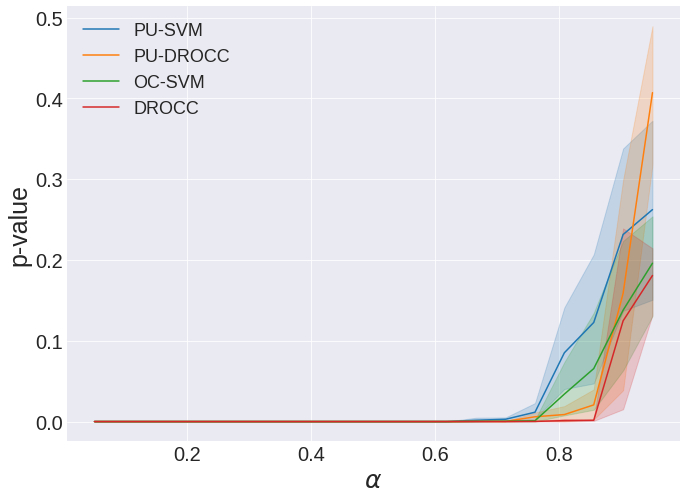}} 
    \subfloat[]{\includegraphics[width=.5\textwidth]{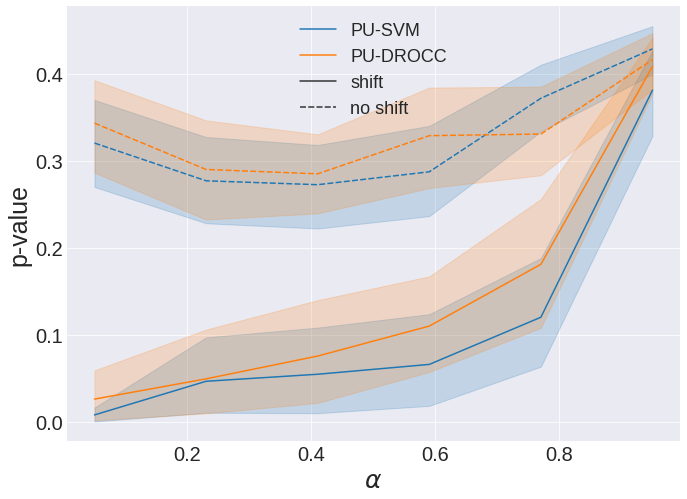}}
\caption{P-values of Mann–Whitney U-test for identification of a) high $\alpha$ or b) negative shift. We show results for OC-SVM and DROCC as OC models and PU-SVM and PU-DROCC as PU models.}
\label{stat_unl}
\end{figure}

As our experiments show, some PU algorithms can perform poorly when available unlabeled data are unreliable. Here, we propose a method to distinguish such cases. Empirically, PU models can struggle when $\alpha$ is too high ($\alpha>0.9$), when negative shifts occur, or when only a few unlabeled examples are at hand. We discover that statistical tests can help with the first two problems. The situation with the high $\alpha$ can be identified by measuring the difference of distributions of outputs of a classification model (OC or PU) on unlabeled and positive samples (fig. \ref{stat_unl} a). The difference can be measured with a statistical test, e.g. T-test or Mann-Whitney U-test. This gives us the following procedure:

\begin{enumerate}
    \item Train an OC or PU model on the available data.

    \item Set the p-value threshold $p_{crit}$ to 0.1 (standard for hypothesis testing).
    
    \item Compute the outputs of the trained model on the samples from positive and unlabeled distributions.
    
    \item Calculate the p-value $pv$ of a statistical test on the predictions from the previous step to check whether the predictions are from the same distribution. If $pv>p_{crit}$ then the null hypothesis of the distributions being the same is accepted and it is advised to use OC models or their robust PU-OC variants, otherwise it is advised to use PU models. 
\end{enumerate}

Similarly, applying Mann-Whitney U-test on the outputs of PU models for train and test unlabeled samples can help to detect distribution shifts. Fig. \ref{stat_unl} (b) shows p-values for samples with and without negative shifts. As can be seen, p-values with shift are significantly lower when $\alpha$ is not high. Therefore, one can detect the shift by comparing p-values of statistical tests on two samples from train distribution and on samples from train and test distribution. The procedure for distribution shift detection is similar to that for high $\alpha$ detection: 

\begin{enumerate}
    \item Train a PU model on the available data. 
  
    \item Calculate the threshold value for p-value $p_{crit}$. Compute the outputs of the trained model on two subsamples from the labeled data. Set $p_{crit}$ as the p-value of the statistical test applied to these subsamples (checking if they come from identical distribution).
    
    \item Compute the outputs of the trained model on the samples from unlabeled train and test distributions.
    
    \item Calculate p-value $pv$ of a statistical test on the predictions from the previous step. If $pv<p_{crit}$ then the distribution shift occurred and it is advised to use OC models or their robust PU-OC variants, otherwise it is advised to use PU models.

\end{enumerate}

\begin{figure}
    \centering
        \includegraphics[width=0.5\textwidth]{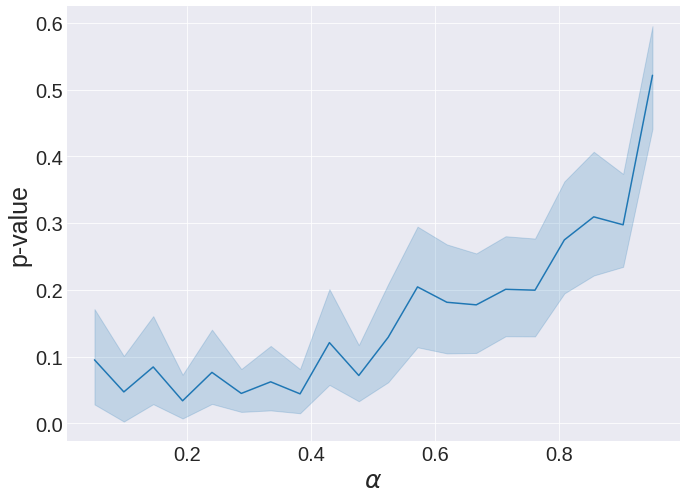}
    \caption{P-values of Mann–Whitney U-test for identification of high $\alpha$ with randomly initialized network.}
    \label{rnd_clf}
\end{figure}

Furthermore, we discovered that even randomly initialized networks can be used for testing if $\alpha$ is high (Fig. \ref{rnd_clf}). However, the resulting statistical test is not as sensitive as the statistical test with trained OC/PU models (Fig. \ref{stat_unl} a). As a remedy, we can use same methodology as in the distribution shift: instead of setting $p_{crit}$ to $0.1$, $p_{crit}$ can be estimated as p-value of the statistical test applied to two subsamples from the positive distribution.

\section{Conclusion}

In this paper, we experimentally compare the effectiveness of state-of-the-art OC and PU algorithms in multiple settings with varied reliability of unlabeled data. We find that while PU algorithms often perform better, their performance can be severely hindered if the unlabeled dataset is too small, contains too few latent negative examples, or is affected by particular shifts of the negative distribution. As an alternative, we propose to incorporate unlabeled data in the pipeline of OC algorithms. We refer to this novel group of hybrid algorithms as PU-OC. We investigate several general ways to construct PU-OC algorithms. We find that PU-OC algorithms can both benefit from reliable unlabeled data and be robust to unreliable unlabeled data. However, this is only the case if they are specifically constructed to have a crucial property: in the absence of latent negative examples in the unlabeled data, the PU-OC algorithm should collapse into the original OC algorithm. Finally, we formulate procedures based on statistical tests that help to identify if the unlabeled data is reliable. Depending on the results of these tests, practitioners can opt to use either state-of-the-art PU algorithms (to benefit from reliable unlabeled data the most) or PU-OC algorithms (that are safe to apply to unreliable unlabeled data). We hope that our findings will motivate future researchers to investigate PU-OC augmentations of their OC algorithms.


To the best of our knowledge, our paper is the first to investigate the potential vulnerability of PU algorithms to the cases of unreliable unlabeled data. While we make a progress towards identifying such cases via statistical tests, a more desirable alternative is a PU algorithm that achieves state-of-the-art performance regardless of whether unlabeled data are reliable. We hope that future researchers will consider testing the robustness of their PU algorithms in the settings of unreliable unlabeled data, as well as proposing safe alternatives to their PU algorithms that should be used in such settings.




\bibliography{ref}
\bibliographystyle{abbrvnat}

\newpage
\appendix

\section{Additional Experiments}\label{app:experiments}

\subsection{PU models}\label{app:experiments_pu}
    Some modern PU methods moved from ROC AUC as the main metric to accuracy. Therefore, it is hard to compare different methods since in each paper they are optimized for their respective metrics. In this experiment, we try to perform a fair comparison between modern and classic PU methods. We choose vehicle classes from CIFAR-10 as positive and animal classes as negatives. The results are reported in Table \ref{pu_sotas}. All methods are described in Appendix \ref{app:pu_methods} and Section \ref{sec:pu_methods}. In Table \ref{pu_sotas}, nnPU row represent nnPU with $\alpha$ estimated from DEDPUL, and nnPU$^{*}$ uses real value of $\alpha$. It can be seen that DRPU outperforms other models when $\alpha <= 0.75$, while PAN outperforms other models when $\alpha >= 0.5$. Since these two models cover the best performance in all cases, we use them as state-of-the-art for comparison in the main text.

\begin{table}
    \caption{ROC AUC of PU models for various $\alpha$}
    \centering
    \label{pu_sotas}
    \begin{tabular}{lccccc}
        \toprule
        $\alpha$& 0.05    & 0.25    & 0.5    & 0.75    & 0.95     \\\hline
        VPU      & \textbf{0.95} & \textbf{0.94} & 0.92 & 0.86 & 0.58  \\
        DRPU     & 0.90 & 0.93 & \textbf{0.93} & \textbf{0.91} & \textbf{0.81}  \\
        PAN      & \textbf{0.95} & \textbf{0.94} & \textbf{0.93} & \textbf{0.91} & 0.69  \\
        EN       & 0.94 & 0.93 & \textbf{0.93} & 0.88 & 0.66\\
        nnPU     & 0.64 & 0.74 & 0.91 & \textbf{0.91} & 0.78\\
        nnPU$^*$ & 0.91 & 0.92 & \textbf{0.93} & 0.85 & 0.72\\
        DEDPUL   & 0.94 & 0.93 & 0.92 & 0.88 & 0.68\\
        \bottomrule
    \end{tabular}
\end{table}

\subsection{OC models}\label{app:experiments_oc}
We perform additional comparison with HRN, another OC model that was published concurrently with CSI. The results are reported in Table \ref{oc_ova}. Despite using the code provided by the authors, HRN performed poorly in our experiments and significantly differs from the results reported in the original paper (reported as HRN$^*$). Our hypothesis to why this could have happened is that, according to the code, the authors have reported the highest ROC-AUC achieved during training, while we evaluate ROC-AUC after the training is complete for all our methods. Still, even if we use the results reported in the original paper for comparison, CSI and DROCC generally outperform HRN, so we do not include it for comparison in the main text.

\begin{table}[h]
    \caption{ROC AUC in one-vs-all setting for OC models}
    \label{oc_ova}
    \centering
    \resizebox{\textwidth}{!}{

    \begin{tabular}{lcccccccccc}
        \toprule
        Pos class& 0    & 1    & 2    & 3    & 4    & 5    & 6   & 7    & 8    & 9    \\\hline
        OC-SVM   & \textbf{0.78} & 0.67 & 0.60 & \textbf{0.67} & 0.63 & 0.64 & 0.74 & 0.66 & 0.68 & 0.75 \\
        CSI    & 0.70 & \textbf{0.87} & 0.63 & 0.62 & 0.60 & \textbf{0.83} & 0.73 & \textbf{0.80} & \textbf{0.86} & \textbf{0.89} \\
        DROCC    & 0.76 & 0.74 & \textbf{0.64} & 0.61 & \textbf{0.71} & 0.67 & \textbf{0.75} & 0.71 & 0.78 & 0.79 \\
        HRN   & 0.74 & 0.5 & 0.5 & 0.5 & 0.5 & 0.5 & 0.5 & 0.5 & 0.67 & 0.55\\
        HRN$^*$   & 0.73 & 0.69 & 0.57 & 0.63 & 0.71 & 0.67 & 0.77 & 0.65 & 0.78 & 0.77\\
        \bottomrule
    \end{tabular}
    }
\end{table}

\section{Methods}\label{app:sec_methods}
\subsection{PU-SVM}\label{app:sec_methods_svm}
This subsection is organized as follows. First, we construct PU-SVM from classic SVM model and describe our implementation that is based on optimization with SGD. Second, we prove that PU-SVM without non-negativity constraint can also be solved via dual problem, albeit less efficiently.

\subsubsection{Our implementation}\label{app:sec_methods_svm_our}
The soft-margin SVM objective can be formulated in the following way: 

\begin{gather}
   \min_{\omega, b} \lambda \norm{w}^2 + E_{(x, y)\sim f_u} l_h(y, w \cdot \Phi(x) - b)
    \label{soft_svm}
\end{gather}

where $l_h(y, x) = max(0, 1 - y x)$ is hinge loss, $\lambda$ is regularization parameter, and $\omega\cdot\Phi(x) - b$ is the decision function. Since the objective (\ref{soft_svm}) is a risk with regularization term, we can apply risk estimation techniques to it. However, several changes are required to that end. First, we replace hinge loss with double hinge loss $l_{dh}(y, x) = \max(-2yx, l_h(y, x))$. As shown in \citep{du2015convex}, the unbiased estimate of the objective (\ref{soft_svm}) with double hinge-loss is a convex function (\ref{upu_svm}).

\begin{gather}
   \min_{\omega, b} \lambda \norm{\omega}^2 + E_{x_u\sim f_u} l_{dh} (-1, \omega \Phi(x_u) - b) +\nonumber\\
    +\alpha E_{x_p\sim f_p} l_{dh} (1, \omega \Phi(x_p) - b) - \alpha E_{x_p\sim f_p} l_{dh} (-1, \omega \Phi(x_p) - b)
    \label{upu_svm}
\end{gather}

As we show in Section \ref{app:sec_methods_svm_dual}, the objective (\ref{upu_svm}) can be solved via dual problem, but the solution is computationally inefficient. As an alternative, we add an additional non-negative constraint to (\ref{upu_svm}) and solve the resulting objective with the stochastic gradient decent (SGD):

\begin{gather}
   \min_{\omega, b} \lambda \norm{\omega}^2 +\alpha E_{x_p\sim f_p} l_{dh} (1, \omega \Phi(x_p) - b)+ \nonumber\\
    +\max(0, E_{x_u\sim f_u} l_{dh} (-1, \omega \Phi(x_u) - b) - \alpha E_{x_p\sim f_p} l_{dh} (-1, \omega \Phi(x_p) - b))
    \label{nnpu_svm}
\end{gather}

The non-negativity constraint is motivated by \citep{kiryo2017positive}. Since we solve (\ref{nnpu_svm}) with SGD, we need to explicitly construct feature space for $x$, i.e. $\Phi(x)$, which again is inefficient. Instead, we make our second change and replace the dot-product in the feature space $w \cdot \Phi(x)$ with the kernel function $K(w, x)$. Note that now $w$ has same dimension as $x$, rather than the dimension of the feature space $\Phi(x)$. Similar trick is applied in OC-NN \citep{chalapathy2018anomaly} and DeepSVDD \citep{ghafoori2020deep}, where the dot-product is replaced with inference of neural network. Finally, we obtain the PU-SVM objective that we use in our study: 

\begin{gather}
   \min_{\omega, b} \lambda \norm{\omega}^2 +\alpha E_{x_p\sim f_p} l_{dh} (1, K(\omega, x) -b)+ \nonumber\\
    +\max(0, E_{x_u\sim f_u} l_{dh} (-1, K(\omega, x) - b) - \alpha E_{x_p\sim f_p} l_{dh} (-1, K(\omega, x) -b))
    \label{re_svm}
\end{gather}

\subsubsection{Dual Problem}\label{app:sec_methods_svm_dual}

Here we show how the objective (\ref{upu_svm}) can be solved via dual problem. To this end, we use the following property of double-hinge loss:
\begin{gather}
    l_{dh}(1, x) - l_{dh}(-1, x) = 
    \max(-2x, 0, 1-x) - \max(2x, 0, 1+x) = -2x
    \label{dhl_prop}
\end{gather}

After replacing the expectations in (\ref{upu_svm}) with their empirical estimates and applying the property (\ref{dhl_prop}), we get the following optimization problem:

\begin{gather}
    \min_{\omega, b} L(\omega, b) = \min_{\omega, b}\left(\lambda \norm{\omega}^2 + \frac 1{n_u} \sum_{i\in u} l_{dh} (-1, \omega \Phi(x_i) - b) 
   + 2\alpha b +  \frac{2
    \alpha}{n_p} \sum_{i\in p} - \omega \Phi(x_i) \right)
    \label{primal}
\end{gather}

Similarly to classic SVM, for each unlabeled point we introduce an additional slack variable $\xi_i = l_{dh} (-1, \omega \Phi(x_i) - b)$, such that:

\begin{gather}
    \xi_i \geq  0 \\
    \xi_i \geq 1 - b + \omega \Phi(x_i) \\
    \xi_i \geq 2 (\omega \Phi(x_i) - b)
\end{gather}

Or, in another form:

\begin{gather}
    -\xi_i \leq 0 \\
    1 - \xi_i - b + \omega \Phi(x_i) \leq 0\\
    -\xi_i - 2 (b - \omega \Phi(x_i)) \leq 0
\end{gather}

According to Karush–Kuhn–Tucker conditions \citep{karush1939minima,kuhn2014nonlinear}, the optimization problem (\ref{primal}) can be solved with the following Lagrangian function:

\begin{gather}
    \mathfrak{L} (\omega, b, \xi, A, B, C) = \lambda \norm{\omega}^2  - \frac{
    \alpha}{n_p} w \sum_{i\in p} \Phi(x_i) 
    +2\alpha b + \frac 1{n_u} \sum_{i\in u} \xi_i - \sum_{i\in u} A_i\xi_i   \nonumber\\
    -\sum_{i\in u} B_i (b - \omega \Phi(x_i) - 1 + \xi_i) 
    -\sum_{i\in u} C_i (\xi_i - 2 (\omega \Phi(x_i) - b))
\end{gather}

As KKT suggests, the optimal vector for the problem above satisfies the following conditions:

\begin{gather}
        A_i\xi_i = 0\\
        B_i (b - \omega \Phi(x_i) - 1 + \xi_i) = 0 \\
        C_i (\xi_i - 2 (\omega \Phi(x_i) - b)) = 0 \\
        \frac{\partial \mathfrak{L}}{\partial \omega} = 
        \frac{\partial \mathfrak{L}}{\partial b} = 
        \frac{\partial \mathfrak{L}}{\partial \xi_i} = 0
\end{gather}

Tacking a closer look at each derivative yields:

\begin{gather}
    \frac{\partial \mathfrak{L}}{\partial \omega} = 2 \omega  - \frac{\alpha}{n_p} \sum_{i\in p} \Phi(x_i) + \sum_{i\in u} (B_i + 2 C_i) \Phi(x_i)\label{dw}\\
    \frac{\partial \mathfrak{L}}{\partial b} = 2\alpha - \sum_{i\in u} B_i - 2\sum_{i \in u} C_i \\
    \frac{\partial \mathfrak{L}}{\partial \xi_i} = \frac{1}{n_u} - A_i - B_i - C_i
\end{gather}

Define $\tau_i$ as:

\begin{gather}
    \tau_i = \begin{cases}
    \frac{\alpha}{2n_p} & ,i\in p\\
    -\frac 12 B_i - C_i & ,i \in u
    \end{cases}
\end{gather}

Then, $\omega$ can be rewritten as:
\begin{gather}
    \omega = \sum_{i\in p, u} \tau_i \Phi(x_i)
    \label{new_w}
\end{gather}

If (\ref{new_w}) is substituted into the Lagrangian, we finally get:
 
\begin{gather}
    \mathfrak{L} = \lambda \sum_{i \in p,u} \sum_{j \in p, u} \tau_i \tau_j \Phi(x_i) \Phi(x_j) - \frac{\alpha}{n_p} \sum_{i \in u, p}\sum_{j \in p} \tau_i \Phi(x_i) \Phi(x_j)\nonumber\\
    +b\underbrace{(2\alpha - \sum_{i\in u} B_i - 2\sum_{i\in u} C_i)}_{\text{0}} 
    + \sum_{i\in u} \xi_i\underbrace{\left(\frac 1{n_u} - A_i - B_i - C_i\right)}_{\text{0}}\nonumber\\
    +\sum_{i\in u} B_i + \sum_{i\in p,u}\sum_{j\in u} \tau_iB_j  \Phi(x_i) \Phi(x_j)  
    +2\sum_{i\in p,u}\sum_{j\in u} \tau_iC_j \Phi(x_i) C_j \Phi(x_j)
    \label{biq_q}
\end{gather}

Subject to:

\begin{gather}
    A_i, B_i, C_i \geq 0 \label{constr_start} \\
    2\alpha - \sum_{i\in u} B_i - 2\sum_{i\in u} C_i = 0\\
    \frac{1}{n_u} - A_i - B_i - C_i = 0 \\
    \tau_i = \begin{cases}
    \frac{\alpha}{2n_p} & ,i\in p\\
    -\frac 12 B_i - C_i & ,i \in u
    \end{cases}
    \label{constr_end}
\end{gather}

We can apply the kernel trick to (\ref{biq_q}) and get a quadratic optimization problem with linear constraints (\ref{constr_start}-\ref{constr_end}):

\begin{gather}
    \mathfrak{L} (\tau, B, C)= \lambda \sum_{i \in p,u} \sum_{j \in p, u} \tau_i \tau_j K(x_i, x_j) - \frac{\alpha}{n_p} \sum_{i \in u, p}\sum_{j \in p} \tau_i K(x_i, x_j)\nonumber\\
    +\sum_{i\in u} B_i + \sum_{i\in p,u}\sum_{j\in u} \tau_i B_j  K(x_i, x_j) + 2\sum_{i\in p,u}\sum_{j\in u} \tau_iC_j K(x_i, x_j)
    \label{final}
\end{gather}

Similarly to the soft-margin SVM, each point with non-zero $\tau_i$ is a support vector. Since, all labeled positive examples have positive $\tau_i$, they are all support vectors. Because of large number of support vectors, this approach is too computationally demanding and time consuming, so we opt for optimization with SGD described in Appendix \ref{app:sec_methods_svm_our}.

\subsection{HRN}
HRN \citep{hu2020hrn} trains a neural classifier on the log-likelihood loss with a special regularization term $\|w\|^{n}$. The proposed term is referred as holistic regularization or H-regularization and essentially is an analogue of lasso or ridge regularization of a higher degree. The paper proposes to set $n=12$. Holistic regularization helps the model to remove feature bias and to prevent it from collapsing into a constant solution. We adapt the official implementation of HRN \footnote{\href{ https://github.com/morning-dews/HRN}{https://github.com/morning-dews/HRN}}.

\subsection{PU methods}\label{app:pu_methods}

\subsubsection{VPU}
    VPU \citep{chen2019vpu} is based on variational inference that allows to estimate a variational upper bound of the KL-divergence between the real posterior distribution and its estimate, parameterized by a neural network. A major advantage of VPU is that variational upper bound can be computed without knowing prior probability $\alpha$, which most modern PU models rely on. Our implementation is based on the pseudo-algorithm provided in the original paper.
    

    

    

\subsubsection{EN}

EN \citep{elkan2008learning} is a classic PU algorithm. It consists of two steps. At the first step, a biased classifier is trained to naively distinguish positive data from unlabeled. At the second step, the output of the trained classifier is calibrated in order to make unbiased predictions.

\subsubsection{DEDPUL}

DEDPUL \citep{ivanov2019dedpul} is a two-stage algorithm. At the first stage, it trains a biased classifier to distinguish positive data from unlabeled and obtains predictions of this classifier for all examples. This is similar to the first stage of EN. At the second stage, it estimates the probability density functions of positive and unlabeled data in the space of predictions. Using the Bayes rule, these densities can estimate the posterior probability $p(x)$, whereas the prior probability $\alpha$ is chosen such that it equals the expected posteriors. DEDPUL achieves state-of-the-art performance in mixture proportion estimation and can improve accuracy of any PU algorithm.

We use the original implementation of DEDPUL\footnote{\href{ https://github.com/dimonenka/DEDPUL}{\label{dedpul}https://github.com/dimonenka/DEDPUL}} without any changes.

\subsection{TIcE}

In all our experiments, we use TIcE estimator of $\alpha$ for PU-SVM and PU-CNN models. We borrow TIcE implementation from DEDPUL repository\hyperref[dedpul]{$^5$}.

\section{Hyperparameters}\label{app:hyperparameters}

Hyperprameters for all models can be found in Tables \ref{hp1}, \ref{hp2}, \ref{hp3}, \ref{hp_pu}, \ref{hp4}, \ref{hp5}. First, we tune hyperparameters on one train-test split via grid search for all models except DROCC. Initial hyperparameters for DROCC we take from \citep{goyal2020drocc}. After that, we manually fine-tune hyperparameters in the neighborhood of the initial parameters found with grid search. For SVM-based methods we search on the following grid: kernel $\in\{rbf, linear\}$, $\nu \in \{0.1, 0.5, 0.9\}$, $\gamma\in \{0.01, 0.1, 1\}$, $\lambda \in\{0.01, 0.1, 1\}$, $lr\in\{10^{-3}, 10^{-4}\}$, num epochs $\in\{50, 100, 200\}$. For LSTM-based methods we search on the following grid: lstm dim $\in\{4, 16\}$, $\nu \in \{0.1, 0.5, 0.9\}$, batch size$\in \{128, 256\}$, $lr\in\{10^{-2}, 10^{-3}, 10^{-4}\}$,num epochs $\in\{50, 100, 200\}$. For CNN-based methods we search on the following grid: $\gamma\in \{0.9, 1\}$, $lr\in\{10^{-3}, 10^{-4}\}$, num epochs $\in\{10, 20, 40\}$.
For CSI-based methods we search on the following grid: $\lambda\in\{0.01, 0.1, 0.5\}$, $lr\in\{10^{-5}, 10^{-4}, 10^{-3}\}$, temperature$\in\{0.01, 0.05, 0.1, 0.5\}$, batch size$\in\{32, 64, 128\}$, $\gamma\in\{0.9, 0.96, 0.99\}$. For DRPU we search on the following grid: $\alpha\in\{0.001, 0.01, 0.1\}$, $lr\in\{10^{-4}, 10^{-3}\}$, $\gamma\in\{0.96, 0.99\}$, num epochs $\in\{10, 20, 50\}$. For PAN we search on the following grid: $\lambda\in\{0.001, 0.01, 0.1\}$, $lr\in\{10^{-4}, 10^{-3}\}$, $\gamma\in\{0.96, 0.99\}$, num epochs $\in\{10, 20, 50\}$. 

\begin{table}[h]
\caption{Hyperparameters for SVM-based models}
\label{hp1}
\centering
\begin{tabular}{lccccccc}
\toprule
Hyperparameter & kernel & $\nu$ & $\gamma$ & $\lambda$ & num epochs & lr & lr decay\\\hline
OC-SVM & linear &0.5 & 2e-3& --& --& --& -- \\
PU-SVM & linear & -- & 1& 0.01& 100& 5e-3& 0.995 \\
\bottomrule
\end{tabular}
\end{table}

\begin{table}[h]
\caption{Hyperparameters for CNN-based models}
\label{hp2}
\centering
\begin{tabular}{lcccc}
\toprule
Hyperparameter & num epochs & lr & batch size & $\gamma$\\\hline
OC-CNN & 10& 1e-4& 256& 1 \\
PU-CNN & 10& 1e-4& 256& 1 \\
\bottomrule
\end{tabular}
\end{table}

\begin{table}[h]
\caption{Hyperparameters for LSTM-based models}
\label{hp3}
\centering
\begin{tabular}{lccccc}
\toprule
Hyperparameter & num epochs & lr & batch size & $\nu$ & lstm dim\\\hline
OC-LSTM & 200& 1e-4& 128& 0.5& 16 \\
PU-LSTM & 100& 5e-3& 256& 0.5& 16 \\
\bottomrule
\end{tabular}
\end{table}

\begin{table}[h]
\caption{Hyperparameters for state-of-the-art PU models}
\label{hp_pu}
\centering
\begin{tabular}{cccccc}
\toprule
Hyperparameter  & $\alpha$ & $\lambda$ & num epochs & lr & $\gamma$\\\hline
DRPU & 0.001& -- &60& 1e-3& 0.99\\
PAN & --& 0.001& 20& 1e-3& 0.99 \\
\bottomrule
\end{tabular}
\end{table}

\begin{table}[h]
    \caption{DROCC Hyperparameters}
    \label{hp4}

    \centering
    \resizebox{\textwidth}{!}{

    \begin{tabular}{lcccccccccccc|c}
        \toprule
        Model & \multicolumn{12}{c|}{DROCC} & PU-DROCC\\\hline
        Pos data& 0    & 1    & 2    & 3    & 4    & 5    & 6   & 7    & 8    & 9 & vehicles &Abnormal & All   \\\hline
        $\lambda$& 0.5    & 0.5    & 0.5    & 0.5    & 0.5    & 0.2    & 0.5   & 0.5    & 0.5    & 0.5 & 0.5 &0.5 & 0.5\\
        radius& 24    & 24    & 32    & 28    & 32    & 36    & 32   & 28    & 28    & 28 & 28 & 30& 2\\
        $\gamma$& 1.5    & 1.1    & 1.5    & 1.1    & 1.5    & 1.5    & 1.5   & 1.1    & 1.1    & 1.1 & 1.1 &1.1 & 2\\
        learning rate& 1e-3    & 1e-3    & 1e-3    & 1e-3    & 1e-3    & 5e-3    & 1e-3   & 1e-3    & 1e-3    & 1e-3 & 1e-3 &1e-3 & 5e-4\\
        ascent step size& 1e-2    & 1e-2    & 1e-2    & 1e-3    & 1e-4    & 1e-3    & 1e-2   & 1e-2    & 1e-2    & 1e-2 & 1e-3 &1e-3&1e-5\\
        ascent num steps& 40    & 60    & 60    & 60    & 20    & 40    & 60   & 50    & 50    & 50 & 60 &50&10 \\
        $\gamma_{lr}$& 1    & 1    & 1    & 1    & 1    & 1    & 1   & 1    & 1    & 1 & 1 &0.99&0.96\\
        num epochs& 30    & 30    & 30    & 30    & 30    & 30    & 30   & 30    & 30    & 30 & 30 &15& 20\\
        batch size& 128    & 128    & 128    & 128    & 128    & 128    & 128   & 128    & 128    & 128 & 128 &128& 256\\
        
        \bottomrule
    \end{tabular}
    }
\end{table}
\begin{table}[h]
    \caption{Hyperparameters for CSI-based models}
    \label{hp5}
    \centering
    \resizebox{\textwidth}{!}{

    \begin{tabular}{lccccccccccccc}
        \toprule
         
         model & Parameter&0 & 1 & 2 & 3 & 4 & 5 & 6 & 7 & 8 & 9 & vehicles & Abnormal \\\hline
            \multirow{6}{*}{CSI} &$\lambda$ & 0.1 & 0.1 & 0.1 & 0.1 & 0.1 & 0.1 & 0.1 & 0.1 & 0.5 & 0.5 & 0.1 & 0.1 \\
            &batch size & 64 & 32 & 32 & 32 & 32 & 32 & 32 & 32 & 32 & 32 & 64 & 64 \\
            &temp & 0.5 &0.5 &0.5 &0.5 &0.5 &0.5 &0.5 &0.5 &0.1 &0.5 &0.5 &0.1 \\
            &$\gamma$ & 0.99 & 0.99 & 0.99 & 0.99 & 0.99 & 0.99 & 0.99 & 0.99 & 0.99 & 0.99 & 0.99 & 0.99 \\

            &$lr$ & 1e-4 & 1e-3 & 1e-3 & 1e-3 & 1e-4 & 1e-3 & 1e-3 & 1e-3 & 1e-3 & 1e-3 & 1e-4 & 1e-3\\
        \bottomrule
        \multirow{6}{*}{PU-CSI} &$\lambda$ & 0.1 & 0.1 & 0.1 & 0.1 & 0.1 & 0.1 & 0.1 & 0.1 & 0.5 & 0.5 & 0.1 & 0.1 \\
            &batch size & 64 &	32 & 64 & 32 & 32 & 32 & 32 & 32 & 32 & 32 & 96 & 64 \\
            &batch size unl & 16 & 16 & 16 & 16 & 16 & 16 & 16 & 16 & 16 & 16 & 32 & 32 \\
            &temp & 0.5&	0.5&	0.5&	0.5	&0.5&	0.5&	0.5	&0.5&	0.1&	0.5	&0.07&	0.01 \\
            &$\gamma$ & 0.99 & 0.99 & 0.99 & 0.99 & 0.99 & 0.99 & 0.99 & 0.99 & 0.99 & 0.99 & 0.99 & 0.99 \\
            &$lr$ & 1e-3 & 1e-3 & 1e-3 & 1e-3 & 1e-4 & 1e-3 & 1e-3 & 1e-3 & 1e-3 & 1e-3 & 5e-3 & 1e-4\\
        \bottomrule
    \end{tabular}
    }
\end{table}
\end{document}